\documentclass[12pt]{article}
\usepackage{arxiv}
\usepackage{authblk}
\usepackage{amsmath, amssymb}
\usepackage{graphicx}
\usepackage{hyperref}
\usepackage[numbers]{natbib}
\usepackage{array}
\usepackage[table]{xcolor}
\usepackage{hhline}
\usepackage{calc}
\usepackage{booktabs}
\usepackage{longtable}
\usepackage{pdflscape}
\usepackage{tabularx}
\usepackage[colorinlistoftodos,prependcaption,textsize=tiny]{todonotes}
\usepackage[title]{appendix}
\usepackage{wasysym} 
\usepackage{tikz}
\usepackage[bottom]{footmisc}
\usepackage[T1]{fontenc}

\usetikzlibrary{arrows.meta,shapes.geometric}

\newcommand{\filledcirc}{\CIRCLE}
\newcommand{\emptycirc}{\Circle}

\title{Persuasion with Large Language Models: A Survey of Empirical Evidence, Study Methodologies, and Ethical Implications}
\renewcommand{\shorttitle}{Persuasion with Large Language Models: A Survey}

\author{Sander Noels}
\author{Alexander Rogiers}
\author{Maarten Buyl}
\author{Tijl De Bie\thanks{Corresponding author. E-mail: \href{mailto:tijl.debie@ugent.be}{tijl.debie@ugent.be}}}
\affil{AIDA-IDLab, Electronics and Information Systems, Ghent University}
\date{}

\begin{document}

\maketitle 
\vspace{-0.5em}
\begin{abstract}
The rapid rise of Large Language Models (LLMs) has created new disruptive possibilities for persuasive communication, enabling fully-automated, personalized, and interactive content generation at an unprecedented scale.
In this paper, we survey the emerging field of \emph{LLM-based persuasion}, reviewing empirical studies that measure the influence of \emph{LLM Systems} on human attitudes and behaviors.
We categorize applications across domains such as politics, marketing, public health, e-commerce, and charitable giving, finding that such systems have frequently achieved human-level or even superhuman persuasiveness.
Synthesizing recent evidence, we identify key factors influencing this effectiveness, including the interaction approach, model scale and capability, prompt design, personalization, and AI source disclosure.
Furthermore, we critically examine the experimental designs and success metrics used to evaluate these Systems, distinguishing between direct behavioral outcomes and proxy indicators.
Our survey suggests that the current capabilities of LLM-based persuasion pose profound ethical and societal risks, including to information integrity, fairness and inclusion, privacy, and individual autonomy.
These risks underscore the urgent need for ethical guidelines and updated regulatory frameworks to avoid the widespread deployment of irresponsible and harmful LLM Systems.
\end{abstract}

\section{Introduction}
\label{sec:intro}

\begin{figure}[h!]
    \centering
    \includegraphics[width=0.85\textwidth]{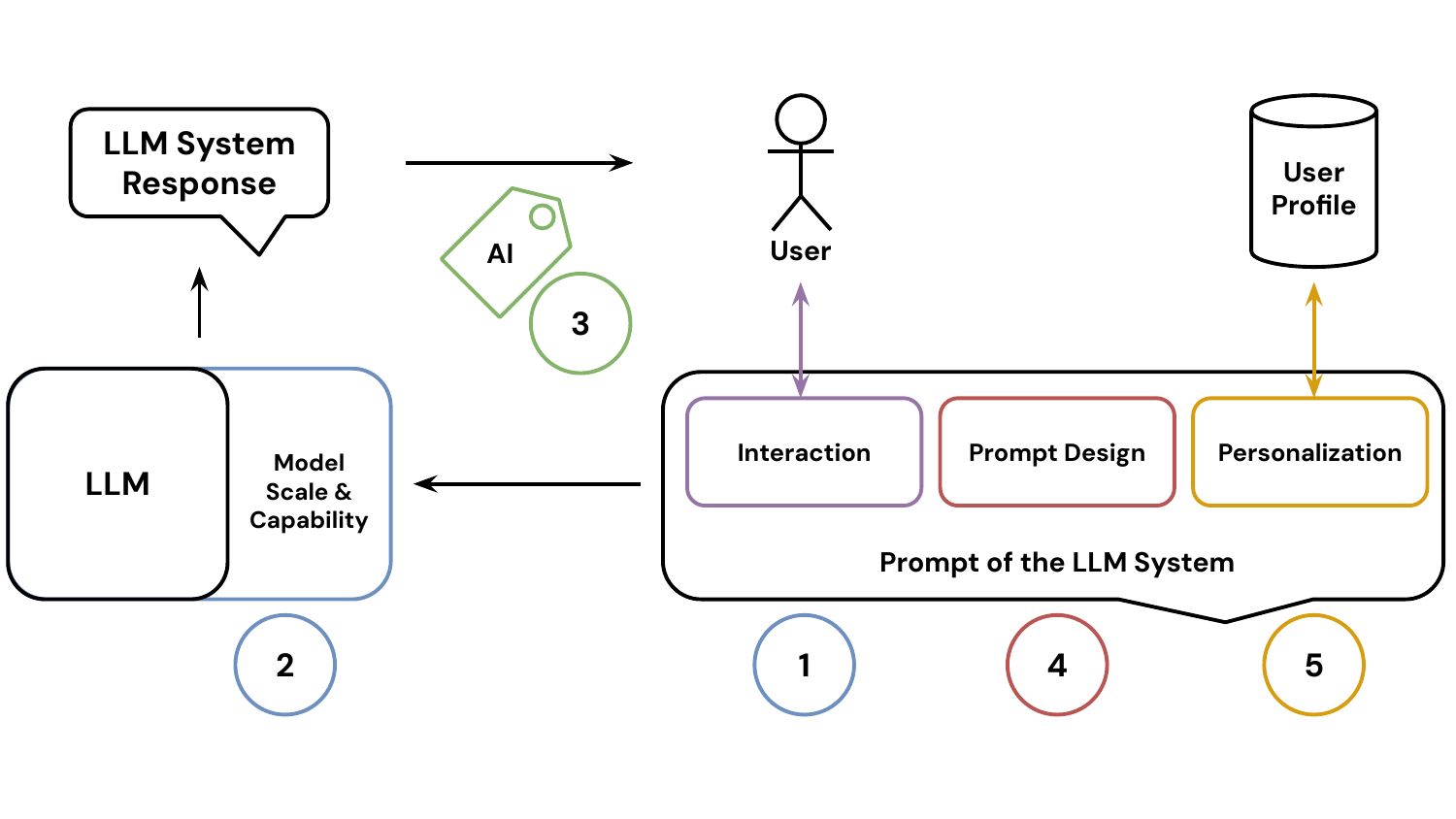} 
    \vspace{-1em}
    \caption{Overview of factors influencing the persuasiveness of an LLM System: (1) the interaction approach (e.g., static vs. interactive), (2) the scale and capabilities of the model, (3) whether AI authorship is disclosed to users, (4) the design of the prompt and associated persuasion strategies, and (5) the extent and method of personalization using user data.}
    \label{fig:factors-overview}
\end{figure}

Persuasive communication, defined as ``any message that is intended to shape, reinforce, or change the responses of another, or others'', has existed both as a practice and as an academic discipline for many decades \citep{stiff2016persuasive}.
Until recently, it has been the prerogative of humans to persuade others, initially through personal or small-group communication, and later through mass-communication channels and technologies such as the printed press, radio, and television.

Since the rise of the internet, persuasion has become increasingly personalized, thus enhancing its reach and effectiveness \citep{guadagno2005online,kaptein2015personalizing,hirsh2012personalized}.
Whereas earlier mass-communication via printed or audio-visual media could be targeted only to a limited extent based on the demographics of a communication channel's audience, the internet allows far more fine-grained personalization, in some applications even down to the individual.

With the advent of Large Language Models (LLMs), marked by the release of ChatGPT in 2022, a new era in persuasive communication has begun.
In just a few years, LLMs have evolved from basic autoregressive text continuation models to models with remarkable emergent capabilities.

We discuss an emergent capability with profound implications: their potential for intentionally generating human-like content that shapes attitudes, informs opinions, and drives behaviors across various domains. 
This evolution enables a significant change in the landscape of persuasion from selecting and presenting human-generated content, albeit possibly in a personalized manner, to automatically generating tailored, context-aware, and hyper-personalized messages at scale.

Several key characteristics make \emph{LLM Systems}, i.e., LLMs integrated with other technologies and deployed for specific applications, particularly effective for persuasive communication.
First of all, they allow for unprecedented \textbf{personalization and adaptability} as they excel at tailoring messages to individual preferences and psychological profiles, creating targeted content that resonates with a targeted group or even a specific individual
.
Second and related to this, LLM Systems appear capable of \textbf{exploiting cognitive biases and heuristics} that make humans vulnerable to persuasion and manipulation, thus allowing them to achieve persuasive impact beyond what was possible with classical scalable approaches
.
Third, they allow for \textbf{interactive conversations}, enabling engaging experiences that can be more effective than traditional one-way messaging
.
Fourth, despite known weaknesses of LLMs in terms of hallucinations, they are capable of a high degree of \textbf{message consistency} even across a prolonged personalized interaction
.
And finally, in comparison to other persuasion methods that share the previously mentioned characteristics, LLM Systems are far more \textbf{scalable} because they can engage in hyper-personalized persuasive interactions at scale without proportional resource increases
.

During the past two years, these key characteristics of LLM Systems have led to the emergence of a new subfield of persuasion research, which we refer to as \emph{LLM-based persuasion}.

The effectiveness of LLM Systems for persuasion depends on various configurable factors, as illustrated in Figure~\ref{fig:factors-overview}. These include the interaction approach (such as whether interactive dialogue is enabled), the scale and capabilities of the language model employed, whether AI authorship is disclosed to users, the design of the prompt, and the extent and method of personalization. Each of these factors can significantly impact the LLM System's persuasive capabilities, as we will discuss in detail in Section~\ref{sec:factors}.

\paragraph{Purpose of this survey.} Through a systematic review of peer-reviewed articles, conference papers, and reputable industry reports published between January 2022 and October 2025, we provide a comprehensive overview of the state-of-the-art in LLM-based persuasion. We examine the applications across various domains of LLM Systems for persuasion, factors and design choices influencing their persuasiveness, the methodological techniques used to study their persuasiveness and thereby quantify progress, and the associated ethical and regulatory challenges. Our survey focuses on experimental studies that directly tested the persuasive capabilities of LLM Systems.

The rest of the paper is organized as follows:
\begin{itemize}
\item Section~\ref{sec:methodology} defines the scope more formally, the inclusion criteria for the surveyed research, and the literature search strategy employed.
\item Section~\ref{sec:application-domains} reviews some of the commonly studied application domains of LLM-based persuasion, including politics, marketing, public health, e-commerce, and efforts to combat misinformation.
\item Section~\ref{sec:factors} analyzes the factors that influence the persuasive capabilities of LLM Systems, such as the interaction approach, model scale and capability, AI source labeling, prompt design, and personalization.
\item Section~\ref{sec:experimental-design} examines the experimental designs employed in LLM persuasion studies, such as rigorous randomized controlled trials, and within- versus between-subject experimental designs. It details structural choices regarding treatment and control conditions, the temporal dimension of measurement, and the emerging use of synthetic simulations.
\item Section~\ref{sec:success-metrics} presents the success metrics used to evaluate persuasive LLM Systems, distinguishing direct persuasion outcomes from proxy indicators of persuasiveness, and summarizes the evidence comparing LLM capabilities to human persuaders.
\item Section~\ref{sec:ethics} discusses the ethical considerations and regulatory landscape surrounding LLM-based persuasion.
\item Section~\ref{sec:conclusion} concludes by summarizing key findings, identifying research gaps, and suggesting directions for future work.
\end{itemize}

By mapping the current state of research on LLM-based persuasion, this survey aims to provide a comprehensive overview of this rapidly evolving field.
In doing so, we hope to raise awareness among researchers, practitioners, policy makers, and the general public of the power of LLM Systems for persuasion today and in the near future,
and to point out the resulting ethical and societal challenges that lie ahead of us.

Table~\ref{tab:papers-overview} provides an overview of the surveyed papers, showing for each paper the application domains it covered, the influencing factors it studied, the experimental design it adopted, the success metrics used, and how persuasive it assessed the investigated LLM System to be when compared to human persuasiveness.

\begin{table}[htbp]
\caption{Summary of papers by Application Domains (Sec.~\ref{sec:application-domains}), Influencing Factors (Sec.~\ref{sec:factors}), Experimental Design (Sec.~\ref{sec:experimental-design}), Success Metrics and Persuasiveness relative to human capabilities (Sec.~\ref{sec:success-metrics}). Filled circles (\filledcirc) indicate the domain/factor/design/metric was explicitly studied in the paper.}
\centering
\tiny
\scalebox{0.9}{
\begin{tabular}{l|ccccc|ccccc|cccccc|cccccc|cccc}
\toprule
& \multicolumn{5}{c|}{\textbf{Application Domains}} & \multicolumn{5}{c|}{\textbf{Influencing Factors}} & \multicolumn{6}{c|}{\textbf{Experimental Design}} & \multicolumn{6}{c|}{\textbf{Success Metrics}} & \multicolumn{4}{c}{\textbf{Persuasiveness}} \\
\rotatebox{90}{\textbf{Citation}} & \rotatebox{90}{\textbf(1) {Public Health}} & \rotatebox{90}{\textbf{(2) Politics}} & \rotatebox{90}{\textbf{(3) E-commerce}} & \rotatebox{90}{\textbf{(4) Misinformation}} & \rotatebox{90}{\textbf{(5) Charity}} & \rotatebox{90}{\textbf{(6) Interaction}} & \rotatebox{90}{\textbf{(7) Model Scale}} & \rotatebox{90}{\textbf{(8) AI Source Labeling}} & \rotatebox{90}{\textbf{(9) Prompt Design}} & \rotatebox{90}{\textbf{(10) Personalization}} & \rotatebox{90}{\textbf{(11) RCT Design}} & \rotatebox{90}{\textbf{(12) Between Subject design}} & \rotatebox{90}{\textbf{(13) Within Subject design}} & \rotatebox{90}{\textbf{(14) Human Control}} & \rotatebox{90}{\textbf{(15) Pre-Post Measure}} & \rotatebox{90}{\textbf{(16) Longitudinal Follow-up}} & \rotatebox{90}{\textbf{(17) Belief Change}} & \rotatebox{90}{\textbf{(18) Attitude Change}} & \rotatebox{90}{\textbf{(19) Behavioral Intent \& Action}} & \rotatebox{90}{\textbf{(20) Source/Misinfo Detection}} & \rotatebox{90}{\textbf{(21) Perceived Effectiveness}} & \rotatebox{90}{\textbf{(22) Text/Comput. Indicators}} & \rotatebox{90}{\textbf{(23) Superhuman}} & \rotatebox{90}{\textbf{(24) On Par}} & \rotatebox{90}{\textbf{(25) Inferior}} & \rotatebox{90}{\textbf{(26) No Comparison}} \\
\midrule

\cite{agarwal2025persuasion} & \emptycirc & \emptycirc & \emptycirc & \filledcirc & \emptycirc & \emptycirc & \filledcirc & \emptycirc & \filledcirc & \emptycirc & \emptycirc & \emptycirc & \filledcirc & \emptycirc & \emptycirc & \emptycirc & \emptycirc & \emptycirc & \emptycirc & \filledcirc & \filledcirc & \filledcirc & \emptycirc & \emptycirc & \emptycirc & \filledcirc \\

\cite{argyle2025testing} & \emptycirc & \filledcirc & \emptycirc & \emptycirc & \emptycirc & \filledcirc & \emptycirc & \emptycirc & \filledcirc & \filledcirc & \filledcirc & \filledcirc & \emptycirc & \emptycirc & \filledcirc & \emptycirc & \emptycirc & \filledcirc & \filledcirc & \filledcirc & \filledcirc & \filledcirc & \emptycirc & \emptycirc & \emptycirc & \filledcirc \\

\cite{ataguba2025persuasion} & \filledcirc & \emptycirc & \emptycirc & \emptycirc & \emptycirc & \emptycirc & \emptycirc & \emptycirc & \emptycirc & \emptycirc & \emptycirc & \emptycirc & \emptycirc & \emptycirc & \emptycirc & \emptycirc & \emptycirc & \emptycirc & \emptycirc & \emptycirc & \filledcirc & \emptycirc & \emptycirc & \emptycirc & \emptycirc & \filledcirc \\

\cite{baiArtificialIntelligenceCan2023} & \emptycirc & \filledcirc & \emptycirc & \emptycirc & \emptycirc & \emptycirc & \emptycirc & \emptycirc & \emptycirc & \emptycirc & \filledcirc & \filledcirc & \emptycirc & \filledcirc & \filledcirc & \emptycirc & \emptycirc & \filledcirc & \emptycirc & \filledcirc & \filledcirc & \filledcirc & \emptycirc & \filledcirc & \emptycirc & \emptycirc \\

\cite{bai2025llm} & \emptycirc & \filledcirc & \emptycirc & \emptycirc & \emptycirc & \emptycirc & \emptycirc & \emptycirc & \emptycirc & \emptycirc & \filledcirc & \filledcirc & \emptycirc & \filledcirc & \filledcirc & \emptycirc & \emptycirc & \filledcirc & \emptycirc & \filledcirc & \filledcirc & \filledcirc & \emptycirc & \filledcirc & \emptycirc & \emptycirc \\

\cite{biswas2025mind} & \emptycirc & \emptycirc & \emptycirc & \emptycirc & \filledcirc & \emptycirc & \emptycirc & \emptycirc & \emptycirc & \emptycirc & \filledcirc & \filledcirc & \emptycirc & \filledcirc & \emptycirc & \emptycirc & \emptycirc & \emptycirc & \filledcirc & \filledcirc & \filledcirc & \emptycirc & \emptycirc & \filledcirc & \emptycirc & \emptycirc \\

\cite{borah2025persuasion} & \emptycirc & \emptycirc & \emptycirc & \filledcirc & \emptycirc & \emptycirc & \filledcirc & \emptycirc & \filledcirc & \emptycirc & \filledcirc & \filledcirc & \emptycirc & \filledcirc & \filledcirc & \emptycirc & \filledcirc & \emptycirc & \emptycirc & \filledcirc & \emptycirc & \filledcirc & \emptycirc & \emptycirc & \emptycirc & \filledcirc \\

\cite{breumPersuasivePowerLarge2024} & \emptycirc & \filledcirc & \emptycirc & \emptycirc & \emptycirc & \emptycirc & \emptycirc & \emptycirc & \filledcirc & \emptycirc & \emptycirc & \emptycirc & \filledcirc & \emptycirc & \emptycirc & \emptycirc & \emptycirc & \emptycirc & \emptycirc & \emptycirc & \filledcirc & \filledcirc & \emptycirc & \emptycirc & \emptycirc & \filledcirc \\

\cite{bohmPeopleDevalueGenerative2023} & \filledcirc & \filledcirc & \emptycirc & \emptycirc & \emptycirc & \emptycirc & \emptycirc & \filledcirc & \emptycirc & \emptycirc & \filledcirc & \filledcirc & \filledcirc & \filledcirc & \emptycirc & \emptycirc & \emptycirc & \emptycirc & \filledcirc & \filledcirc & \filledcirc & \emptycirc & \emptycirc & \filledcirc & \emptycirc & \emptycirc \\

\cite{calle2024towards} & \filledcirc & \emptycirc & \emptycirc & \emptycirc & \emptycirc & \emptycirc & \filledcirc & \emptycirc & \filledcirc & \emptycirc & \emptycirc & \emptycirc & \filledcirc & \filledcirc & \emptycirc & \emptycirc & \emptycirc & \emptycirc & \emptycirc & \emptycirc & \filledcirc & \filledcirc & \emptycirc & \filledcirc & \emptycirc & \emptycirc \\

\cite{carrasco-farreLargeLanguageModels2024} & \emptycirc & \filledcirc & \emptycirc & \emptycirc & \emptycirc & \emptycirc & \emptycirc & \emptycirc & \filledcirc & \emptycirc & \emptycirc & \filledcirc & \emptycirc & \filledcirc & \filledcirc & \emptycirc & \emptycirc & \filledcirc & \emptycirc & \emptycirc & \emptycirc & \filledcirc & \emptycirc & \filledcirc & \emptycirc & \emptycirc \\

\cite{chenWouldAIChatbot2023} & \emptycirc & \emptycirc & \filledcirc & \emptycirc & \emptycirc & \emptycirc & \emptycirc & \emptycirc & \emptycirc & \emptycirc & \emptycirc & \emptycirc & \emptycirc & \emptycirc & \emptycirc & \emptycirc & \emptycirc & \emptycirc & \filledcirc & \emptycirc & \filledcirc & \emptycirc & \emptycirc & \emptycirc & \emptycirc & \filledcirc \\

\cite{chen2025framework} & \emptycirc & \filledcirc & \emptycirc & \emptycirc & \emptycirc & \emptycirc & \emptycirc & \filledcirc & \emptycirc & \emptycirc & \filledcirc & \filledcirc & \emptycirc & \filledcirc & \filledcirc & \filledcirc & \emptycirc & \filledcirc & \emptycirc & \emptycirc & \emptycirc & \emptycirc & \emptycirc & \filledcirc & \emptycirc & \emptycirc \\

\cite{cheng2025towards} & \emptycirc & \emptycirc & \emptycirc & \emptycirc & \emptycirc & \filledcirc & \filledcirc & \emptycirc & \emptycirc & \emptycirc & \emptycirc & \emptycirc & \filledcirc & \emptycirc & \filledcirc & \emptycirc & \filledcirc & \filledcirc & \emptycirc & \emptycirc & \emptycirc & \filledcirc & \emptycirc & \emptycirc & \emptycirc & \filledcirc \\

\cite{cima2025contextualized} & \emptycirc & \filledcirc & \emptycirc & \emptycirc & \emptycirc & \emptycirc & \emptycirc & \emptycirc & \filledcirc & \filledcirc & \emptycirc & \filledcirc & \filledcirc & \emptycirc & \emptycirc & \emptycirc & \emptycirc & \emptycirc & \emptycirc & \emptycirc & \filledcirc & \filledcirc & \emptycirc & \emptycirc & \emptycirc & \filledcirc \\

\cite{coppolillo2025engagement} & \emptycirc & \filledcirc & \emptycirc & \emptycirc & \emptycirc & \emptycirc & \filledcirc & \emptycirc & \emptycirc & \filledcirc & \emptycirc & \emptycirc & \emptycirc & \filledcirc & \emptycirc & \emptycirc & \emptycirc & \emptycirc & \emptycirc & \emptycirc & \emptycirc & \filledcirc & \emptycirc & \filledcirc & \emptycirc & \emptycirc \\

\cite{corro2025exploring} & \filledcirc & \emptycirc & \emptycirc & \emptycirc & \emptycirc & \emptycirc & \filledcirc & \emptycirc & \emptycirc & \emptycirc & \emptycirc & \emptycirc & \emptycirc & \emptycirc & \emptycirc & \emptycirc & \emptycirc & \emptycirc & \emptycirc & \emptycirc & \filledcirc & \emptycirc & \emptycirc & \emptycirc & \emptycirc & \filledcirc \\

\cite{costelloDurablyReducingConspiracy2024} & \emptycirc & \emptycirc & \emptycirc & \filledcirc & \emptycirc & \emptycirc & \emptycirc & \emptycirc & \emptycirc & \emptycirc & \filledcirc & \filledcirc & \emptycirc & \emptycirc & \filledcirc & \filledcirc & \filledcirc & \emptycirc & \filledcirc & \emptycirc & \emptycirc & \emptycirc & \emptycirc & \emptycirc & \emptycirc & \filledcirc \\

\cite{dash2025persuasive} & \filledcirc & \filledcirc & \emptycirc & \emptycirc & \emptycirc & \emptycirc & \emptycirc & \emptycirc & \emptycirc & \emptycirc & \filledcirc & \filledcirc & \emptycirc & \filledcirc & \emptycirc & \emptycirc & \filledcirc & \emptycirc & \filledcirc & \filledcirc & \emptycirc & \filledcirc & \emptycirc & \filledcirc & \emptycirc & \emptycirc \\

\cite{doudkin2025synthetic} & \emptycirc & \filledcirc & \emptycirc & \emptycirc & \emptycirc & \filledcirc & \emptycirc & \emptycirc & \filledcirc & \filledcirc & \filledcirc & \filledcirc & \emptycirc & \emptycirc & \filledcirc & \emptycirc & \filledcirc & \filledcirc & \filledcirc & \emptycirc & \emptycirc & \emptycirc & \emptycirc & \emptycirc & \emptycirc & \filledcirc \\

\cite{durmus2024persuasion} & \emptycirc & \filledcirc & \emptycirc & \emptycirc & \emptycirc & \emptycirc & \filledcirc & \emptycirc & \filledcirc & \emptycirc & \emptycirc & \filledcirc & \emptycirc & \filledcirc & \filledcirc & \emptycirc & \emptycirc & \filledcirc & \emptycirc & \emptycirc & \emptycirc & \filledcirc & \emptycirc & \filledcirc & \emptycirc & \emptycirc \\

\cite{el2024improving} & \emptycirc & \filledcirc & \emptycirc & \emptycirc & \emptycirc & \emptycirc & \filledcirc & \emptycirc & \filledcirc & \filledcirc & \emptycirc & \emptycirc & \filledcirc & \filledcirc & \emptycirc & \emptycirc & \emptycirc & \emptycirc & \emptycirc & \emptycirc & \filledcirc & \filledcirc & \filledcirc & \emptycirc & \emptycirc & \emptycirc \\

\cite{elaraby2024persuasiveness} & \emptycirc & \emptycirc & \emptycirc & \emptycirc & \emptycirc & \emptycirc & \filledcirc & \emptycirc & \filledcirc & \emptycirc & \emptycirc & \emptycirc & \filledcirc & \emptycirc & \emptycirc & \emptycirc & \emptycirc & \emptycirc & \emptycirc & \emptycirc & \filledcirc & \filledcirc & \emptycirc & \emptycirc & \emptycirc & \filledcirc \\

\cite{elhissoufi2024leveraging} & \emptycirc & \emptycirc & \filledcirc & \emptycirc & \emptycirc & \emptycirc & \emptycirc & \emptycirc & \emptycirc & \emptycirc & \emptycirc & \emptycirc & \emptycirc & \emptycirc & \emptycirc & \emptycirc & \emptycirc & \emptycirc & \emptycirc & \emptycirc & \filledcirc & \filledcirc & \emptycirc & \emptycirc & \emptycirc & \filledcirc \\

\cite{fetrati2025leveraging} & \filledcirc & \emptycirc & \emptycirc & \emptycirc & \emptycirc & \emptycirc & \emptycirc & \emptycirc & \emptycirc & \emptycirc & \emptycirc & \emptycirc & \emptycirc & \emptycirc & \emptycirc & \emptycirc & \emptycirc & \emptycirc & \emptycirc & \emptycirc & \filledcirc & \emptycirc & \emptycirc & \emptycirc & \emptycirc & \filledcirc \\

\cite{furumaiZeroshotPersuasiveChatbots2024} & \filledcirc & \emptycirc & \filledcirc & \emptycirc & \filledcirc & \emptycirc & \filledcirc & \emptycirc & \emptycirc & \emptycirc & \emptycirc & \filledcirc & \emptycirc & \emptycirc & \emptycirc & \emptycirc & \emptycirc & \emptycirc & \emptycirc & \filledcirc & \filledcirc & \filledcirc & \emptycirc & \emptycirc & \emptycirc & \filledcirc \\

\cite{goldsteinHowPersuasiveAIgenerated2024} & \emptycirc & \filledcirc & \emptycirc & \emptycirc & \emptycirc & \emptycirc & \emptycirc & \emptycirc & \filledcirc & \emptycirc & \filledcirc & \filledcirc & \filledcirc & \filledcirc & \emptycirc & \emptycirc & \filledcirc & \filledcirc & \emptycirc & \emptycirc & \filledcirc & \emptycirc & \emptycirc & \filledcirc & \emptycirc & \emptycirc \\

%
\cite{hackenburgComparingPersuasivenessRoleplaying2023} & \filledcirc & \filledcirc & \emptycirc & \emptycirc & \emptycirc & \emptycirc & \emptycirc & \emptycirc & \filledcirc & \filledcirc & \filledcirc & \filledcirc & \emptycirc & \filledcirc & \emptycirc & \emptycirc & \emptycirc & \filledcirc & \emptycirc & \filledcirc & \emptycirc & \emptycirc & \filledcirc & \emptycirc & \emptycirc & \emptycirc \\

\cite{hackenburgEvaluatingPersuasiveInfluence2024} & \emptycirc & \filledcirc & \emptycirc & \emptycirc & \emptycirc & \emptycirc & \emptycirc & \emptycirc & \filledcirc & \filledcirc & \filledcirc & \filledcirc & \emptycirc & \emptycirc & \emptycirc & \emptycirc & \emptycirc & \filledcirc & \emptycirc & \filledcirc & \emptycirc & \emptycirc & \emptycirc & \emptycirc & \emptycirc & \filledcirc \\

\cite{hackenburg2025levers} & \emptycirc & \filledcirc & \emptycirc & \emptycirc & \emptycirc & \filledcirc & \filledcirc & \emptycirc & \filledcirc & \filledcirc & \filledcirc & \filledcirc & \emptycirc & \emptycirc & \filledcirc & \filledcirc & \emptycirc & \filledcirc & \emptycirc & \emptycirc & \emptycirc & \filledcirc & \emptycirc & \emptycirc & \emptycirc & \filledcirc \\

\cite{hackenburg2025scaling} & \emptycirc & \filledcirc & \emptycirc & \emptycirc & \emptycirc & \emptycirc & \filledcirc & \emptycirc & \emptycirc & \emptycirc & \filledcirc & \filledcirc & \emptycirc & \filledcirc & \emptycirc & \emptycirc & \emptycirc & \filledcirc & \emptycirc & \filledcirc & \emptycirc & \filledcirc & \emptycirc & \filledcirc & \emptycirc & \emptycirc \\

\cite{hackenburgEvidenceLogScaling2024} & \emptycirc & \filledcirc & \emptycirc & \emptycirc & \emptycirc & \emptycirc & \filledcirc & \emptycirc & \emptycirc & \emptycirc & \filledcirc & \filledcirc & \emptycirc & \filledcirc & \emptycirc & \emptycirc & \emptycirc & \filledcirc & \emptycirc & \emptycirc & \emptycirc & \filledcirc & \emptycirc & \filledcirc & \emptycirc & \emptycirc \\

\cite{havin2025can} & \emptycirc & \filledcirc & \emptycirc & \emptycirc & \emptycirc & \filledcirc & \emptycirc & \emptycirc & \emptycirc & \emptycirc & \filledcirc & \filledcirc & \emptycirc & \filledcirc & \filledcirc & \emptycirc & \emptycirc & \filledcirc & \emptycirc & \emptycirc & \emptycirc & \emptycirc & \emptycirc & \filledcirc & \emptycirc & \emptycirc \\

\cite{he2025enhancing} & \filledcirc & \filledcirc & \emptycirc & \emptycirc & \emptycirc & \emptycirc & \filledcirc & \emptycirc & \filledcirc & \emptycirc & \emptycirc & \emptycirc & \emptycirc & \emptycirc & \emptycirc & \emptycirc & \emptycirc & \emptycirc & \emptycirc & \emptycirc & \filledcirc & \filledcirc & \emptycirc & \emptycirc & \emptycirc & \filledcirc \\

\cite{jin2024persuading} & \filledcirc & \filledcirc & \filledcirc & \emptycirc & \filledcirc & \emptycirc & \filledcirc & \emptycirc & \filledcirc & \emptycirc & \emptycirc & \emptycirc & \filledcirc & \filledcirc & \emptycirc & \emptycirc & \emptycirc & \emptycirc & \emptycirc & \emptycirc & \filledcirc & \filledcirc & \filledcirc & \emptycirc & \emptycirc & \emptycirc \\

\cite{karakacs2025changes} & \filledcirc & \emptycirc & \emptycirc & \emptycirc & \emptycirc & \emptycirc & \emptycirc & \emptycirc & \emptycirc & \emptycirc & \emptycirc & \emptycirc & \filledcirc & \emptycirc & \filledcirc & \emptycirc & \filledcirc & \filledcirc & \emptycirc & \emptycirc & \filledcirc & \filledcirc & \emptycirc & \emptycirc & \emptycirc & \filledcirc \\

\cite{karinshakWorkingAIPersuade2023a} & \filledcirc & \emptycirc & \emptycirc & \emptycirc & \emptycirc & \emptycirc & \emptycirc & \filledcirc & \filledcirc & \emptycirc & \filledcirc & \filledcirc & \emptycirc & \filledcirc & \emptycirc & \emptycirc & \emptycirc & \emptycirc & \emptycirc & \emptycirc & \filledcirc & \filledcirc & \filledcirc & \emptycirc & \emptycirc & \emptycirc \\

\cite{kong2025huper} & \emptycirc & \emptycirc & \emptycirc & \emptycirc & \emptycirc & \emptycirc & \filledcirc & \emptycirc & \filledcirc & \filledcirc & \emptycirc & \emptycirc & \filledcirc & \filledcirc & \emptycirc & \emptycirc & \emptycirc & \emptycirc & \emptycirc & \emptycirc & \filledcirc & \filledcirc & \filledcirc & \emptycirc & \emptycirc & \emptycirc \\

\cite{kumar2025large} & \filledcirc & \emptycirc & \emptycirc & \emptycirc & \emptycirc  & \filledcirc & \emptycirc & \emptycirc & \filledcirc & \emptycirc  & \filledcirc & \filledcirc & \emptycirc & \emptycirc & \filledcirc & \filledcirc  & \emptycirc & \emptycirc & \filledcirc & \emptycirc & \emptycirc & \emptycirc  & \emptycirc & \emptycirc & \emptycirc & \filledcirc \\

%
\cite{lim2023artificial} & \filledcirc & \emptycirc & \emptycirc & \emptycirc & \emptycirc & \emptycirc & \emptycirc & \emptycirc & \emptycirc & \emptycirc & \emptycirc & \emptycirc & \filledcirc & \filledcirc & \emptycirc & \emptycirc & \emptycirc & \emptycirc & \emptycirc & \emptycirc & \filledcirc & \filledcirc & \filledcirc & \emptycirc & \emptycirc & \emptycirc \\

\cite{limEffectSourceDisclosure2024} & \filledcirc & \emptycirc & \emptycirc & \emptycirc & \emptycirc & \emptycirc & \emptycirc & \filledcirc & \emptycirc & \emptycirc & \filledcirc & \filledcirc & \filledcirc & \filledcirc & \emptycirc & \emptycirc & \emptycirc & \emptycirc & \emptycirc & \emptycirc & \filledcirc & \filledcirc & \emptycirc & \filledcirc & \emptycirc & \emptycirc \\

\cite{lin2025persuading} & \emptycirc & \filledcirc & \emptycirc & \emptycirc & \emptycirc & \emptycirc & \filledcirc & \emptycirc & \filledcirc & \filledcirc & \filledcirc & \filledcirc & \emptycirc & \emptycirc & \filledcirc & \filledcirc & \emptycirc & \filledcirc & \filledcirc & \emptycirc & \filledcirc & \filledcirc & \emptycirc & \emptycirc & \emptycirc & \filledcirc \\

\cite{matzPotentialGenerativeAI2024} & \filledcirc & \filledcirc & \filledcirc & \emptycirc & \emptycirc & \emptycirc & \emptycirc & \filledcirc & \filledcirc & \filledcirc & \filledcirc & \filledcirc & \filledcirc & \emptycirc & \emptycirc & \emptycirc & \emptycirc & \emptycirc & \filledcirc & \emptycirc & \filledcirc & \emptycirc & \emptycirc & \emptycirc & \emptycirc & \filledcirc \\

\cite{meguellati2025duality} & \emptycirc & \filledcirc & \filledcirc & \emptycirc & \emptycirc& \emptycirc & \emptycirc & \emptycirc & \emptycirc & \filledcirc& \filledcirc & \filledcirc & \filledcirc & \filledcirc & \emptycirc & \emptycirc& \emptycirc & \emptycirc & \filledcirc & \emptycirc & \filledcirc & \emptycirc& \emptycirc & \filledcirc & \emptycirc & \emptycirc \\

\cite{meguellatiHowGoodAre2024} & \emptycirc & \emptycirc & \filledcirc & \emptycirc & \emptycirc & \emptycirc & \emptycirc & \emptycirc & \filledcirc & \filledcirc & \emptycirc & \filledcirc & \filledcirc & \filledcirc & \emptycirc & \emptycirc & \emptycirc & \emptycirc & \filledcirc & \emptycirc & \filledcirc & \emptycirc & \emptycirc & \filledcirc & \emptycirc & \emptycirc \\

\cite{metzgerEmpoweringCalibratedDis2024} & \emptycirc & \emptycirc & \emptycirc & \emptycirc & \emptycirc & \emptycirc & \emptycirc & \emptycirc & \filledcirc & \emptycirc & \filledcirc & \filledcirc & \emptycirc & \emptycirc & \filledcirc & \emptycirc & \emptycirc & \filledcirc & \emptycirc & \emptycirc & \filledcirc & \emptycirc & \emptycirc & \emptycirc & \emptycirc & \filledcirc \\

\cite{nezhad2025adaptive} & \emptycirc & \emptycirc & \emptycirc & \emptycirc & \filledcirc & \emptycirc & \emptycirc & \emptycirc & \filledcirc & \filledcirc & \emptycirc & \filledcirc & \emptycirc & \emptycirc & \emptycirc & \emptycirc & \emptycirc & \emptycirc & \filledcirc & \emptycirc & \filledcirc & \emptycirc & \emptycirc & \emptycirc & \emptycirc & \filledcirc \\

\cite{palmerLargeLanguageModels2023} & \emptycirc & \filledcirc & \emptycirc & \emptycirc & \emptycirc & \emptycirc & \emptycirc & \filledcirc & \emptycirc & \emptycirc & \filledcirc & \filledcirc & \emptycirc & \filledcirc & \emptycirc & \emptycirc & \emptycirc & \emptycirc & \emptycirc & \emptycirc & \filledcirc & \filledcirc & \emptycirc & \filledcirc & \emptycirc & \emptycirc \\

\cite{pauliMeasuringBenchmarkingLarge2024} & \emptycirc & \filledcirc & \emptycirc & \emptycirc & \filledcirc & \emptycirc & \filledcirc & \emptycirc & \filledcirc & \emptycirc & \emptycirc & \emptycirc & \filledcirc & \filledcirc & \emptycirc & \emptycirc & \emptycirc & \emptycirc & \emptycirc & \emptycirc & \filledcirc & \filledcirc & \emptycirc & \emptycirc & \emptycirc & \filledcirc \\

\cite{potter2024hidden} & \emptycirc & \filledcirc & \emptycirc & \emptycirc & \emptycirc & \emptycirc & \filledcirc & \emptycirc & \emptycirc & \emptycirc & \filledcirc & \filledcirc & \emptycirc & \emptycirc & \filledcirc & \emptycirc & \emptycirc & \filledcirc & \filledcirc & \emptycirc & \filledcirc & \filledcirc & \emptycirc & \emptycirc & \emptycirc & \filledcirc \\

\cite{qin2024beyond} & \emptycirc & \emptycirc & \filledcirc & \emptycirc & \emptycirc & \emptycirc & \emptycirc & \emptycirc & \filledcirc & \emptycirc & \emptycirc & \emptycirc & \filledcirc & \emptycirc & \filledcirc & \emptycirc & \emptycirc & \emptycirc & \filledcirc & \emptycirc & \filledcirc & \filledcirc & \emptycirc & \emptycirc & \emptycirc & \filledcirc \\

\cite{saenger2024autopersuade} & \emptycirc & \emptycirc & \emptycirc & \emptycirc & \emptycirc & \emptycirc & \emptycirc & \emptycirc & \filledcirc & \emptycirc & \emptycirc & \emptycirc & \filledcirc & \emptycirc & \emptycirc & \emptycirc & \emptycirc & \emptycirc & \emptycirc & \emptycirc & \filledcirc & \filledcirc & \emptycirc & \filledcirc & \emptycirc & \emptycirc \\

\cite{salvi2025conversational} & \emptycirc & \filledcirc & \emptycirc & \emptycirc & \emptycirc & \emptycirc & \emptycirc & \emptycirc & \emptycirc & \filledcirc & \filledcirc & \filledcirc & \emptycirc & \filledcirc & \filledcirc & \emptycirc & \emptycirc & \filledcirc & \emptycirc & \filledcirc & \emptycirc & \filledcirc & \filledcirc & \emptycirc & \emptycirc & \emptycirc \\


\cite{sasaki2025ai} & \filledcirc & \emptycirc & \emptycirc & \emptycirc & \emptycirc & \filledcirc & \emptycirc & \emptycirc & \filledcirc & \emptycirc & \filledcirc & \filledcirc & \emptycirc & \emptycirc & \filledcirc & \emptycirc & \emptycirc & \filledcirc & \emptycirc & \emptycirc & \filledcirc & \filledcirc & \emptycirc & \emptycirc & \emptycirc & \filledcirc \\

\cite{schneiders2025objection} & \emptycirc & \emptycirc & \emptycirc & \emptycirc & \emptycirc & \emptycirc & \emptycirc & \filledcirc & \emptycirc & \emptycirc & \filledcirc & \filledcirc & \filledcirc & \filledcirc & \emptycirc & \emptycirc & \emptycirc & \emptycirc & \filledcirc & \filledcirc & \emptycirc & \filledcirc & \filledcirc & \emptycirc & \emptycirc & \emptycirc \\

\cite{schoenegger2025large} & \emptycirc & \emptycirc & \emptycirc & \filledcirc & \emptycirc & \emptycirc & \emptycirc & \emptycirc & \emptycirc & \emptycirc & \filledcirc & \filledcirc & \filledcirc & \filledcirc & \emptycirc & \emptycirc & \filledcirc & \emptycirc & \filledcirc & \filledcirc & \emptycirc & \filledcirc & \filledcirc & \emptycirc & \emptycirc & \emptycirc \\

\cite{shin2025adoption} & \emptycirc & \emptycirc & \filledcirc & \emptycirc & \emptycirc & \emptycirc & \emptycirc & \emptycirc & \filledcirc & \emptycirc & \filledcirc & \emptycirc & \filledcirc & \filledcirc & \emptycirc & \emptycirc & \emptycirc & \emptycirc & \filledcirc & \emptycirc & \filledcirc & \filledcirc & \filledcirc & \emptycirc & \emptycirc & \emptycirc \\

\cite{simchonPersuasiveEffectsPolitical2024} & \emptycirc & \filledcirc & \emptycirc & \emptycirc & \emptycirc & \emptycirc & \filledcirc & \emptycirc & \filledcirc & \filledcirc & \filledcirc & \filledcirc & \emptycirc & \emptycirc & \emptycirc & \emptycirc & \emptycirc & \emptycirc & \emptycirc & \emptycirc & \filledcirc & \filledcirc & \emptycirc & \emptycirc & \emptycirc & \filledcirc \\    

\cite{spitaleAIModelGPT32023} & \filledcirc & \emptycirc & \emptycirc & \filledcirc & \emptycirc & \emptycirc & \emptycirc & \emptycirc & \emptycirc & \emptycirc & \emptycirc & \emptycirc & \filledcirc & \filledcirc & \filledcirc & \emptycirc & \emptycirc & \emptycirc & \emptycirc & \filledcirc & \emptycirc & \emptycirc & \filledcirc & \emptycirc & \emptycirc & \emptycirc \\

\cite{sun2026cutting} & \emptycirc & \emptycirc & \filledcirc & \emptycirc & \emptycirc & \emptycirc & \emptycirc & \filledcirc & \filledcirc & \emptycirc & \filledcirc & \filledcirc & \filledcirc & \filledcirc & \emptycirc & \emptycirc & \emptycirc & \emptycirc & \emptycirc & \emptycirc & \filledcirc & \emptycirc & \filledcirc & \emptycirc & \emptycirc & \emptycirc \\

\cite{teigenPersuasivenessArgumentsAIsource2024} & \filledcirc & \filledcirc & \emptycirc & \emptycirc & \emptycirc & \emptycirc & \emptycirc & \filledcirc & \filledcirc & \emptycirc & \filledcirc & \filledcirc & \emptycirc & \emptycirc & \emptycirc & \emptycirc & \emptycirc & \emptycirc & \emptycirc & \emptycirc & \filledcirc & \emptycirc & \emptycirc & \emptycirc & \emptycirc & \filledcirc \\

\cite{timm2025tailored} & \emptycirc & \filledcirc & \emptycirc & \emptycirc & \emptycirc & \filledcirc & \emptycirc & \emptycirc & \filledcirc & \filledcirc & \emptycirc & \emptycirc & \filledcirc & \filledcirc & \filledcirc & \emptycirc & \emptycirc & \filledcirc & \emptycirc & \emptycirc & \emptycirc & \emptycirc & \filledcirc & \emptycirc & \emptycirc & \emptycirc \\

\cite{vahidov2025customer} & \emptycirc & \emptycirc & \filledcirc & \emptycirc & \emptycirc & \emptycirc & \emptycirc & \emptycirc & \emptycirc & \emptycirc & \filledcirc & \filledcirc & \emptycirc & \emptycirc & \emptycirc & \emptycirc & \emptycirc & \emptycirc & \filledcirc & \filledcirc & \filledcirc & \emptycirc & \emptycirc & \emptycirc & \emptycirc & \filledcirc \\

\cite{wilczynski2024resistance} & \emptycirc & \emptycirc & \emptycirc & \filledcirc & \emptycirc & \emptycirc & \filledcirc & \emptycirc & \filledcirc & \emptycirc & \emptycirc & \emptycirc & \filledcirc & \emptycirc & \filledcirc & \emptycirc & \emptycirc & \emptycirc & \filledcirc & \filledcirc & \emptycirc & \filledcirc & \emptycirc & \emptycirc & \emptycirc & \filledcirc \\

\cite{wu2024mindshift} & \filledcirc & \emptycirc & \emptycirc & \emptycirc & \emptycirc & \emptycirc & \emptycirc & \emptycirc & \filledcirc & \filledcirc & \emptycirc & \emptycirc & \filledcirc & \emptycirc & \filledcirc & \emptycirc & \filledcirc & \filledcirc & \filledcirc & \emptycirc & \filledcirc & \emptycirc & \emptycirc & \emptycirc & \emptycirc & \filledcirc \\

\cite{xia2025comparison} & \filledcirc & \emptycirc & \emptycirc & \emptycirc & \emptycirc & \emptycirc & \emptycirc & \emptycirc & \filledcirc & \emptycirc & \emptycirc & \emptycirc & \filledcirc & \filledcirc & \emptycirc & \emptycirc & \emptycirc & \emptycirc & \emptycirc & \emptycirc & \filledcirc & \emptycirc & \filledcirc & \emptycirc & \emptycirc & \emptycirc \\

\cite{yoonDesigningEvaluatingMultiChatbot2024} & \emptycirc & \emptycirc & \emptycirc & \emptycirc & \filledcirc & \emptycirc & \emptycirc & \emptycirc & \emptycirc & \emptycirc & \emptycirc & \emptycirc & \filledcirc & \emptycirc & \emptycirc & \emptycirc & \emptycirc & \emptycirc & \filledcirc & \emptycirc & \filledcirc & \emptycirc & \emptycirc & \emptycirc & \emptycirc & \filledcirc \\

\cite{zhangHumanFavoritismNot2023} & \filledcirc & \emptycirc & \filledcirc & \emptycirc & \emptycirc & \emptycirc & \emptycirc & \filledcirc & \emptycirc & \emptycirc & \filledcirc & \filledcirc & \emptycirc & \filledcirc & \emptycirc & \emptycirc & \emptycirc & \emptycirc & \emptycirc & \emptycirc & \filledcirc & \emptycirc & \filledcirc & \emptycirc & \emptycirc & \emptycirc \\

\cite{zhou2025communication} & \filledcirc & \emptycirc & \emptycirc & \filledcirc & \emptycirc & \emptycirc & \filledcirc & \emptycirc & \emptycirc & \emptycirc & \emptycirc & \emptycirc & \filledcirc & \filledcirc & \emptycirc & \emptycirc & \emptycirc & \emptycirc & \emptycirc & \filledcirc & \filledcirc & \filledcirc & \filledcirc & \emptycirc & \emptycirc & \emptycirc \\

\cite{zhouSyntheticLiesUnderstanding2023} & \filledcirc & \emptycirc & \emptycirc & \filledcirc & \emptycirc & \emptycirc & \emptycirc & \emptycirc & \filledcirc & \emptycirc & \emptycirc & \emptycirc & \emptycirc & \filledcirc & \emptycirc & \emptycirc & \emptycirc & \emptycirc & \emptycirc & \filledcirc & \emptycirc & \filledcirc & \emptycirc & \emptycirc & \emptycirc & \filledcirc \\

\bottomrule
\end{tabular}
}
\label{tab:papers-overview}
\end{table}

\section{Methodology}\label{sec:methodology}

\paragraph{Scope}
We aim to consider literature that measures the persuasiveness of \textit{LLM Systems}, i.e., systems that deliver LLM-generated text to humans through chatbots, social media bots, or other human-facing interfaces. While Section~\ref{sec:intro} provided a broad conceptual definition of persuasion, for the purposes of this survey we operationalize \textit{persuasion} specifically as an intentional attempt to change a person's attitudes, beliefs, or behaviour in a specified direction, operationalized via human outcomes such as attitude or opinion change, choice or compliance, or scores on a validated persuasion-related scale.

\paragraph{Eligibility criteria}
We \emph{include} papers that satisfy all of the following criteria:
\begin{enumerate}
    \item They use an LLM to generate text with the intention to persuade;
    \item report a human-facing outcome (e.g., attitude/opinion shift, conversion/compliance, or a validated persuasion scale) with sufficient methodological detail to assess the study design;
    \item are first made public between \textbf{1 Jan 2022} and \textbf{30 Oct 2025};
    \item are indexed by Google Scholar;
    \item are written in English.
\end{enumerate}

We start from 2022 because this period marks the emergence and widespread deployment of modern conversational LLMs, most notably with the release of ChatGPT in November 2022~\cite{openai_introducing_2022}. We further include non–peer-reviewed preprints that meet the above criteria, as the field is rapidly evolving and many relevant studies appear first on preprint servers.

The criteria are intended to \textit{exclude} papers that (a) are theoretical or opinion-only pieces; (b) only measure persuasion using intrinsic text properties (e.g., linguistic features) without a recipient response; (c) only employ algorithms that recommend, but do not generate, text (such as social media ranking algorithms); (d) only measure effects adjacent to persuasion, such as detecting whether text is AI-generated; or (e) focus on physically embodied LLM Systems, such as hardware social robots.

\paragraph{Search strategy}
We queried Scopus on 30 Oct 2025 with the following search string:
\begin{align*}
    &\textit{(``large language model'' OR ``LLM'' OR ``generative AI'')}\\
    &\textit{AND (``persuasion'' OR ``persuasive'' OR ``debate'' OR ``persuasiveness'')}\\
    &\textit{AND (``human'' OR ``user study'' OR ``randomized control trial'')}\\
    &\textit{AND PUBYEAR > 2021 AND PUBYEAR < 2027}\\
    &\textit{AND ( LIMIT-TO ( LANGUAGE , ``English'' ) )}
\end{align*}
This query returned $4{,}783$ peer-reviewed articles (journal and conference papers) and $4{,}171$ preprints (e.g., on arXiv). Results were sorted by Scopus' built-in relevance score. We then manually screened titles and abstracts of the top $200$ peer-reviewed articles and the top $50$ preprints, stopping at these cut-offs because the yield of potentially relevant studies dropped sharply in the lower-ranked results. During screening, we removed duplicates within and across the two sets; when both a preprint and a peer-reviewed version of the same work were present, we retained only the peer-reviewed version. The remaining unique studies that satisfied our eligibility criteria constitute the corpus synthesized in this survey.

\paragraph{Synthesis}
For each included paper, we extracted structured notes on the aspects most relevant to this survey: application domain, influencing factors (including the direction and magnitude of persuasive effects), methodological setup (e.g., experimental design, sample, and comparison conditions), success metrics, and overall conclusions regarding persuasiveness. We then used these annotations to organize the literature along the dimensions discussed in the remainder of the paper, most notably the application domains in Section~\ref{sec:application-domains} and the methodological and outcome-focused analyses in later sections.

\section{Application Domains}
\label{sec:application-domains}

LLM Systems have been deployed across various domains to influence human behavior and attitudes. This section examines five commonly studied application areas, each presenting unique challenges and opportunities for persuasive AI.

\subsection{Public Health}

Public health applications (see \emph{column (1) in Table~\ref{tab:papers-overview}}) uniquely combine the need for scientific accuracy with persuasive impact. Research in this domain has investigated LLM Systems ranging from automated message generators for social media to fully interactive therapeutic agents.

\paragraph{Generating public health messages}
A significant portion of research focuses on generating public health messaging for mass communication. In the context of vaccination, \citet{karinshakWorkingAIPersuade2023a} deployed GPT-3 to draft COVID-19 appeals, comparing them against official CDC messaging. This application extends to HPV vaccination, where \citet{xia2025comparison} used GPT-4 to address sensitive sub-topics like side effects and stigma. Beyond vaccines, \citet{lim2023artificial} utilized the open-source Bloom model to generate tweets about prenatal health (specifically folic acid), while \citet{limEffectSourceDisclosure2024} examined the deployment of AI-generated health messages under different source disclosure conditions.

\paragraph{Addiction and cessation assistants}
Moving from general broadcasting to targeted interventions, several studies have designed systems to assist with addiction and cessation. \citet{calle2024towards} investigated the use of LLMs to generate smoking cessation intervention messages that meet clinical standards for smartphone delivery. Addressing digital well-being, \citet{wu2024mindshift} developed `MindShift,' a context-aware Android application that uses LLMs to analyze user states and generate real-time interventions for problematic smartphone usage. 

\paragraph{Promoting healthy lifestyles}
LLM Systems have also been applied to modify lifestyle and dietary behaviors through various interaction modes. \citet{karakacs2025changes} utilized open-ended conversational interfaces to reduce psychological commitment to meat consumption. Taking a different approach, \citet{sasaki2025ai} investigated indirect persuasion regarding nutritional advice by having users observe a peer AI accept advice, thereby triggering a conformity effect. \citet{matzPotentialGenerativeAI2024} further explored how such lifestyle messaging—across health and other domains—can be personalized to individual psychological profiles.

\paragraph{Coaching and counselling}
Finally, researchers are exploring the use of LLMs for sensitive subjects and prolonged health and wellness engagement. \citet{fetrati2025leveraging} utilized LLM-based counseling to promote STI prevention among young adults, deploying AI to discuss sensitive topics like HIV transmission. In the realm of mental wellness, \citet{kumar2025large} designed agents with sociable personas to improve engagement with mindfulness interventions over several weeks. Expanding the scope to embodied interaction, \citet{corro2025exploring} developed an LLM-controlled virtual coach that monitors physiological signals to guide users through breathing exercises in real-time. In these advisory roles, \citet{bohmPeopleDevalueGenerative2023} investigated how users perceive the competence of LLM Systems providing advice on personal and societal challenges.

\subsection{Politics}

Political communication is a primary arena for LLM-based persuasion (see \emph{column (2) in Table~\ref{tab:papers-overview}}), where systems are deployed to generate policy appeals, campaign materials, propaganda, and interactive debates.

\paragraph{Policy Statements and Legislative Arguments}
A foundational use case involves the generation of static arguments supporting specific legislative measures or policy positions. \citet{baiArtificialIntelligenceCan2023} and \citet{bai2025llm} demonstrated that LLMs could write short arguments supporting measures such as assault weapon bans and carbon taxes, effectively shifting participants' opinions. Similarly, \citet{argyle2025testing} used LLMs to influence policy attitudes on divisive issues like immigration and education, while concurrently measuring impacts on voting intention. The quality of these AI-generated arguments often rivals human output; \citet{palmerLargeLanguageModels2023} found LLM-written arguments on U.S. politics to be generally easier to read and more positive in tone than human versions, while \citet{durmus2024persuasion} showed that advanced models like Claude 3 Opus produce arguments on emerging policy issues with persuasiveness similar to human-written ones. While \citet{teigenPersuasivenessArgumentsAIsource2024} note that labeling these arguments as AI-generated can reduce their persuasive impact, the capability remains robust. \citet{hackenburgEvidenceLogScaling2024} further quantified this in a large-scale study, finding that a single static LLM message shifted policy attitudes by about 6 percentage points compared to controls
.

\paragraph{Political Advertising and Campaign Messaging}
Beyond policy essays, LLM Systems are increasingly used to generate and optimize campaign advertisements. In direct comparisons with human experts, \citet{hackenburgComparingPersuasivenessRoleplaying2023} found that GPT-4’s political messages were about as persuasive as those of professional consultants, and often more effective on right-leaning issues. Research has also focused on the potential for microtargeting these ads. \citet{simchonPersuasiveEffectsPolitical2024} found that personality-matched political ads were judged to be more persuasive than generic ones and that such personalization can be scaled automatically. However, results on the effectiveness of microtargeting are mixed; \citet{hackenburgEvaluatingPersuasiveInfluence2024} found that while GPT-4 messages were broadly persuasive, microtargeted versions were not necessarily more effective than high-quality generic messages.
LLMs are also employed to analyze and optimize the reach of campaign content. \citet{meguellati2025duality} applied an LLM-based detector to election ads and found that ads classified as “highly persuasive” attracted substantially more spending and impressions. Furthermore, \citet{coppolillo2025engagement} demonstrated that LLM agents can be fine-tuned to maximize content engagement within polarized communities; using data from the Brexit referendum, they showed that even smaller, specialized models could generate highly viral content adapted to the dominant sentiment of a social network.

\paragraph{Propaganda and Disinformation}
The ability to generate persuasive text at scale has raised concerns about the automation of propaganda and disinformation. \citet{goldsteinHowPersuasiveAIgenerated2024} showed that GPT-3-generated propaganda articles were highly persuasive and nearly as effective as human-written propaganda on the same topics. Highlighting a more specific risk, \citet{timm2025tailored} demonstrated that "threat actor" agents could be instructed to hallucinate plausible-sounding statistics to support their arguments. When these fabricated stats were deployed in personalized debates, the AI was significantly more successful at shifting user stances than standard human arguments, raising concerns about the feasibility of automated, large-scale influence operations.

\paragraph{Interactive Debates and Voter Engagement}
Moving beyond static content, recent research investigated the use of LLMs for interactive political dialogue and debating. In live multi-turn debates, \citet{salvi2025conversational} found that a GPT-4 agent was about as persuasive as a human opponent, and significantly outperformed humans when given minimal personal information about the interlocutor. These effects appear to translate to real-world electoral contexts. Focusing on the 2024 U.S. presidential election, \citet{potter2024hidden} found that interactive dialogues shifted candidate preference, with nearly 20\% of Trump supporters reducing their support after the interaction. Similarly, \citet{lin2025persuading} conducted pre-registered survey experiments across three countries, finding that three-round AI dialogues significantly shifted voter preferences and policy support with effects partially persisting for over a month. However, \citet{chen2025framework} highlight the practical constraints of these interactive tools, estimating that while LLM chatbots are as persuasive as human video ads on a per-person basis, the high cost of incentivizing voters to actually engage with a chatbot currently limits their scalability compared to traditional broadcast media.

\paragraph{Depolarization and Counterspeech}
Finally, LLM Systems are being explored as tools to improve political discourse rather than just persuade. Taking an approach focused on depolarization, \citet{el2024improving} utilized LLMs to rewrite "ineffective" arguments into versions specifically styled to resonate with liberal or conservative worldviews, finding that these ideologically-tuned rewrites were rated significantly more effective than the originals. In the context of online toxicity, \citet{cima2025contextualized} used LLMs to generate counterspeech in political Reddit communities; they found that responses contextualized using conversation and user history were rated significantly more likely to persuade toxic users to re-engage in civil discourse than generic AI responses.

\subsection{E-commerce and Marketing}

In marketing and e-commerce (see \emph{column (3) in Table~\ref{tab:papers-overview}}), LLM Systems are deployed to drive consumer decisions and business outcomes. Research has investigated specific use cases ranging from interactive shopping assistants and active sales agents to the generation of advertising content and the management of consumer complaints.

\paragraph{Shopping Chatbots and Recommendations}
A primary application is the use of conversational agents to recommend products and services. \citet{chenWouldAIChatbot2023} showed that specific features of an AI shopping chatbot’s recommendations increased users’ intention to adopt them, primarily by fostering cognitive and emotional trust. To enhance the reliability of such systems, \citet{furumaiZeroshotPersuasiveChatbots2024} developed a zero-shot chatbot for travel recommendations that grounds its suggestions in retrieved facts, improving persuasiveness while maintaining factual accuracy. Similarly, \citet{qin2024beyond} designed a conversational movie recommender that balances persuasiveness with credibility, ensuring that explanations induce higher watching intentions without relying on hallucinations or deceptive claims.

\paragraph{Sales and Negotiations}
Beyond standard recommendations, LLM agents are capable of handling complex negotiation and sales tasks. \citet{vahidov2025customer} demonstrated that LLM-powered agents negotiating phone plans achieved significantly higher prices than pre-scripted agents. Crucially, this increased yield did not alienate customers; the LLM agent was rated as fairer and resulted in higher intentions to return. LLMs can also support human sales professionals; \citet{elhissoufi2024leveraging} showed that GPT-4 could convert technical product specifications into structured sales arguments that experts rated as highly coherent and aligned with customer motives.

\paragraph{Ads and Campaigns}
In an early study, \citet{zhangHumanFavoritismNot2023} observed that ChatGPT-4-generated ads were judged to be of higher quality than those created by human experts, although the gap reduced when revealing AI or human authorship.
LLMs also function as scalable copywriters capable of high degrees of personalization. \citet{meguellatiHowGoodAre2024} and \citet{meguellati2025duality} found that GPT-generated personalized ads performed comparably to human-written ones. \citet{matzPotentialGenerativeAI2024} demonstrated that personality-tailored ads generated by ChatGPT were more persuasive than non-personalized versions, sometimes increasing willingness to pay. 

\paragraph{Complaints Handling}
Finally, LLMs are being applied to post-purchase interactions. \citet{shin2025adoption} found that LLMs can refine consumer complaints to be clearer and more professional, making them more likely to result in relief.

\subsection{Mitigating and Generating Misinformation}

Research in this domain investigates two opposing applications: the generation of deceptive content and the deployment of AI to detect or debunk it (see \emph{column (4) in Table~\ref{tab:papers-overview}}).

\paragraph{Generating misinformation}
On the adversarial side, studies have modeled how LLMs can be used to generate misinformation that mimics human credibility cues. \citet{zhouSyntheticLiesUnderstanding2023} examined the creation of AI-generated misinformation, noting e.g. its tendency for enhancing details and including cognitive processing expressions, communicating uncertainties, and simulating personal tones and emotions, thus enhancing its credibility. \citet{spitaleAIModelGPT32023} tested the generation of social media disinformation, finding that readers could not reliably distinguish it from human-authored content. Beyond static text, \citet{schoenegger2025large} investigated the use of LLMs to actively steer users toward incorrect answers in `cognitive illusion' tasks, demonstrating a use case where the system significantly reduced participant accuracy compared to control conditions.

\paragraph{Mitigating misinformation}
In terms of mitigation, researchers have explored various specific interventions to counter falsehoods. One prominent application is interactive debunking: \citet{costelloDurablyReducingConspiracy2024} deployed GPT-4 Turbo to engage conspiracy believers in personalized, multi-turn dialogues, a method that successfully reduced belief in specific conspiracies. Another application is automated fact-checking, where the specific style of the AI is scrutinized. \citet{zhou2025communication} investigated the generation of fact-checks and rebuttals, finding that the ``neutral'' and ``objective'' tone naturally adopted by LLMs can be an asset, as readers perceived these AI rebuttals as less accusatory and easier to follow than human-written ones.

\paragraph{Detecting manipulative behavior}
Finally, systems have been designed for the technical detection of manipulation. For instance, \citet{wilczynski2024resistance} proposed the use of LLMs as `Manipulation Fuses'—classifiers designed to verify outputs—showing that models can be effectively deployed to detect and flag manipulative statements within a given context.

\subsection{Charity}

Charitable giving and prosocial campaigns (see \emph{column (5) in Table~\ref{tab:papers-overview}}) pose unique persuasive tasks for LLM Systems: motivating generosity without overstepping into manipulation. Research in this domain has primarily distinguished between two main use cases: the use of LLM Systems for rewriting and fine-tuning persuasive messages, and the deployment of fully interactive persuasive chatbots for soliciting charitable donations.

\paragraph{Writing campaign materials}
The first major use case involves using LLMs as drafting assistants to rewrite or co-write static campaign materials. \citet{pauliMeasuringBenchmarkingLarge2024} investigated the capability of LLMs to rewrite short texts, including donation appeals, to modulate their persuasiveness level while preserving the core message. Similarly, \citet{biswas2025mind} examined the use of multilingual LLM co-writing tools to assist human writers in crafting persuasive fundraising advertisements for organizations like the World Wildlife Fund.

\paragraph{Soliciting donations}
In the context of interactive chatbots, researchers have investigated how conversational agents can solicit donations through dialogue. \citet{yoonDesigningEvaluatingMultiChatbot2024} explored multi-agent interfaces, designing a system where two GPT-based chatbots represented Save the Children and UNICEF to promote donations interactively. Building on this conversational approach, \citet{furumaiZeroshotPersuasiveChatbots2024} introduced PersuaBot, a multi-step LLM and retrieval pipeline designed to generate factual, persuasive appeals specifically for donation solicitation. Other work has focused on the internal mechanisms of such chatbots; \citet{nezhad2025adaptive} introduced a framework that dynamically switches between specific persuasive strategies (e.g., logical vs. emotional appeals) based on real-time sentiment analysis, investigating whether responsive adaptation allows agents to better navigate the donation conversation compared to static strategies.
Finally, \citet{jin2024persuading} studied the potential of transfer learning in this context, demonstrating that their PersuGPT model trained on diverse daily persuasion tasks was also effective in the charity domain, evaluated on the \textit{PersuasionForGood} benchmark \citep{wang2019persuasion}.

\subsection{Other Application Domains}

Beyond the major categories discussed above, LLMs are being investigated in specialized verticals ranging from professional services to environmental advocacy, as well as in domain-agnostic decision support. Here we briefly discuss a few additional examples.

In the \emph{professional services sector}, systems are being designed for high-stakes persuasive interactions. In the \emph{financial domain}, \citet{kong2025huper} applied reinforcement learning to the adversarial task of debt collection, optimizing dialogue models to persuade users to repay overdue debts. In the \emph{legal field}, \citet{schneiders2025objection} explored the application of LLMs for generating lay legal advice, investigating their persuasive influence across the subdomains of traffic, planning, and property law.

Researchers are also exploring whether LLMs can \emph{drive pro-environmental behaviors and sustainable lifestyle changes}. \citet{doudkin2025synthetic} compared personalized chatbots to static text for encouraging sustainable daily habits, such as reducing single-use plastics. Specific dietary changes linked to sustainability have also been a focus; \citet{saenger2024autopersuade} developed a framework to generate and evaluate arguments for veganism, while \citet{karakacs2025changes} investigated whether brief conversations with ChatGPT-4o could influence beliefs regarding the necessity of meat consumption.

Finally, some researchers are evaluating LLMs as \emph{domain-agnostic tools for debate assistance and general knowledge formation}. Unlike targeted campaigns, these applications aim to help users navigate complex topics or update their beliefs. \citet{elaraby2024persuasiveness} examined the use of LLMs for selecting the best argument from two given options, and generating a persuasive rationale for the choice made. In the realm of general knowledge, \citet{schoenegger2025large} utilized a quiz-based framework to test AI persuasion on trivia, cognitive illusions, and future event predictions, studying the system's ability to steer participants toward specific answers.

\section{Factors Influencing Persuasiveness}
\label{sec:factors}

The effectiveness of LLM-driven persuasion is shaped by a complex interplay of factors. This section examines five key elements identified in recent research that influence the persuasive capabilities of LLM Systems: the interaction approach, model scale and capability, AI source labeling, prompt design, and personalization. Each factor represents a different dimension of how LLMs can be designed or deployed to sway opinions or behavior. Note that Table~\ref{tab:papers-overview} indicates only which studies experimentally vary a factor to measure its persuasive effect.

\subsection{Interaction Approach}

The degree of interactivity (see \emph{column (6) in Table~\ref{tab:papers-overview}}) in an LLM’s communication with users---ranging from one-shot message delivery to multi-turn dialogues or debates---can markedly affect persuasive impact.

Many studies have used one-way, static message exposure in their designs: participants simply read an LLM-generated message with no back-and-forth exchange. For instance, \citet{baiArtificialIntelligenceCan2023} had subjects read a persuasive political statement generated by an LLM (GPT-3/3.5) and measured attitude change pre- vs. post-exposure, demonstrating that even a single-shot AI message can shift opinions. Likewise, \citet{simchonPersuasiveEffectsPolitical2024} and \citet{hackenburgEvaluatingPersuasiveInfluence2024} each evaluated persuasion by showing participants tailored political advertisements or essays and recording their immediate effect; while these messages were sometimes microtargeted to user attributes, the communication remained non-interactive. Such passive delivery approaches, including personalized ads in marketing contexts \citep{meguellatiHowGoodAre2024}, dominate the current literature due to the relative ease of testing one-shot persuasion. Variations within static delivery have also been tested; for instance, \citet{bai2025llm} included a ``human-in-the-loop'' condition where participants selected the most persuasive AI draft to show to others. Interestingly, this curated approach was not statistically more effective than the raw LLM output, suggesting that human selection does not necessarily improve upon automated generation.

At the other end of the spectrum, research has explored interactive, multi-turn conversations and debates. In these settings, the LLM System engages the user (or another agent) in a dialogue, adapting its persuasive strategy based on the interlocutor’s responses. For example, \citet{costelloDurablyReducingConspiracy2024} deployed GPT-4 in a three-round conversation to debunk individuals’ conspiracy beliefs; the AI dynamically injected participant-specific content into each response, leading to about a 21\% reduction in belief adherence that persisted weeks later. Similarly, \citet{salvi2025conversational} examined persuasion during live multi-turn debates between humans and an AI, finding that conversational dynamics significantly influenced the AI’s success. Other studies have implemented chatbot-style interactions. \citet{furumaiZeroshotPersuasiveChatbots2024} had users chat with an LLM through several simulated turns, and \citet{yoonDesigningEvaluatingMultiChatbot2024} introduced multiple AI agents in a chatroom to present information interactively. Expanding on these multi-agent dynamics, \citet{sasaki2025ai} introduced a ``peer'' AI into a group chat; they found that users showed significantly greater attitude change when they observed this peer agent being persuaded by the main AI, successfully leveraging the human tendency for social conformity. These interactive approaches intuitively allow LLMs to address user doubts or counterarguments in real time, a capability hypothesized by researchers such as \citet{hackenburgEvaluatingPersuasiveInfluence2024} and \citet{durmus2024persuasion} to enhance persuasion on complex topics.

While early work often focused on either static or interactive modes in isolation, recent experiments have directly compared the two, yielding mixed results regarding the inherent superiority of dialogue. Supporting the interaction hypothesis, \citet{hackenburg2025levers} compared static messages against interactive dialogue within the same large-scale experiment and found that while static messages were effective, the conversational AI was substantially more persuasive. However, other studies contradict this finding. \citet{havin2025can} 
found that static paragraphs were just as effective at shifting opinions as real-time conversation. Similarly, \citet{argyle2025testing} randomly assigned participants to receive either a static message or engage in a six-turn chat (using direct argumentation or motivational interviewing) and found that interactive conditions were no more effective at changing policy attitudes than a single, static generic message.

These discrepancies suggest that the specific design of the interactivity and the nature of the outcome measure are critical. For instance, \citet{kumar2025large} found that adding a sociable, interactive agent to static video content increased user engagement, whereas a reflective questioning agent did not. Furthermore, when the bar is raised to behavioral change rather than attitude shift, the mode of delivery may matter less; \citet{doudkin2025synthetic} compared static statements against interactive chatbots for environmental advocacy and found that neither approach yielded significant self-reported behavioral changes.

In summary, 
static messages offer simplicity and proven effectiveness, while interactive dialogues provide opportunities for adaptation that can—under the right conditions—drive stronger persuasion, though they are not a guaranteed silver bullet.

\subsection{Model Scale and Capability}

Model scale (see \emph{column (7) in Table~\ref{tab:papers-overview}}) is a fundamental factor influencing persuasive capability. In general, larger models tend to demonstrate more advanced language understanding, coherence, and reasoning, which can translate into more convincing arguments. Multiple studies have observed that as LLMs increase in scale, their persuasive power grows accordingly. For instance, \citet{durmus2024persuasion} found a clear scaling trend in persuasiveness: newer, more capable models (e.g., Claude 3) produced arguments nearly as effective at changing opinions as human-crafted ones, whereas smaller models (e.g., the compact Claude Instant 1.2) had significantly less impact. Similarly, \citet{simchonPersuasiveEffectsPolitical2024} reported that ChatGPT (based on a more advanced GPT-3.5-tier model) achieved a stronger persuasive effect in a political microtargeting task than its predecessor GPT-3. This trend holds for open-source models as well: \citet{elaraby2024persuasiveness} found that Llama-2-70B-chat produced highly persuasive rationales, outperforming its 7B and 13B variants. Qualitative differences in performance underpin these quantitative gains. \citet{pauliMeasuringBenchmarkingLarge2024} observed that the smallest model (7B parameters) was less effective at changing its persuasive language when instructed, whereas the larger models more faithfully followed instructions to modulate their persuasiveness. 

However, the relationship between model size and persuasive power is not linear. Evidence suggests diminishing returns once models reach very large scales. In a massive between-model study, \citet{hackenburgEvidenceLogScaling2024} deployed a range of LLMs from roughly 70 million parameters up to state-of-the-art models like GPT-4, and found that increases in persuasiveness followed a sublinear pattern (approximately logarithmic) with respect to model size. Beyond a certain threshold, additional model capacity yielded only modest gains in persuasive effectiveness.

This implies that while scaling up an LLM greatly boosts its basic language and reasoning abilities, beyond a point other factors such as training data quality or specialized fine-tuning may play a larger role in further enhancing persuasiveness. Supporting this nuance, \citet{hackenburg2025levers} found that specific post-training techniques could allow a smaller open-source model to match or outperform the persuasive capability of a much larger frontier model. Even more strikingly, \citet{coppolillo2025engagement} found that a small, 2B parameter model (Gemma-2B) fine-tuned specifically for engagement maximization could outperform significantly larger state-of-the-art models, including LLaMA-3.1-70B and ChatGPT-4o, in generating content that propagates through a social network.

In summary, while model scale is a crucial determinant of an LLM’s persuasive prowess, specialized training strategies can allow smaller models to achieve comparable or even superior influence.

\subsection{AI Source Labeling}

AI source labeling (see \emph{column (8) in Table~\ref{tab:papers-overview}}) refers to whether the audience is told an AI system wrote the message (e.g., labeled “AI-generated” vs. presented as human-written or with no attribution). Many studies have manipulated this factor by giving identical content different source labels to test its influence on persuasion. A common finding is that content explicitly labeled as AI-authored is often received with more skepticism and lower persuasiveness than the same content attributed to a human.

For example, \citet{bohmPeopleDevalueGenerative2023} observed that revealing AI authorship led participants to rate the source as less competent, even though the advice’s quality and their intention to use or share it were unchanged. Similarly, \citet{karinshakWorkingAIPersuade2023a} found that while a GPT-3 system could generate highly effective vaccine messages, participants expressed a clear dispreference for those messages when they were labeled as coming from an AI rather than from human authorities like the CDC. In the political domain, \citet{palmerLargeLanguageModels2023} noted a slight drop in readers’ preference for arguments when an “AI algorithm” was disclosed as the author.
In the context of product and campaign ads, \citet{zhangHumanFavoritismNot2023} observed that while ChatGPT-4-generated ads were judged to be of higher quality than those created by human experts, revealing the authorship triggered a ``human favoritism'' bias, boosting the perceived quality of the human-created ads and thereby reducing (though not reversing) the perceived quality gap.
\citet{teigenPersuasivenessArgumentsAIsource2024} provided further evidence: in a controlled experiment across multiple topics, arguments labeled as AI-authored were judged significantly less persuasive than identical arguments labeled as human-produced (especially when the human source was described as an expert). This finding echoes results in the legal domain, where \citet{schneiders2025objection} observed a ``superhuman'' effect in which participants were more willing to act on LLM-generated advice than lawyer advice when the source was hidden. However, this advantage evaporated when the source was labeled, demonstrating that AI disclosure can nullify performance gaps favoring the LLM System.

Not all studies find a strong labeling penalty. For example, \citet{matzPotentialGenerativeAI2024} report that any negative effect of AI disclosure is minimal when messages are personality-tailored to recipients. In their experiments, AI-generated, trait-matched messages were rated as similarly persuasive even when AI authorship was disclosed, raising the question whether personalization might blunt disclosure penalties.

\subsection{Prompt Design}

Prompt design—the deliberate wording and context given to the LLM—significantly impacts the system’s persuasive effectiveness (see \emph{column (9) in Table~\ref{tab:papers-overview}}). By adjusting how a request or instruction is framed, one can shape the tone, style, and content of the AI’s output. This includes specifying a role or persona for the LLM (e.g., an expert or authoritative figure), setting the desired tone (formal, empathetic, assertive), choosing a rhetorical framing (such as narrative storytelling versus citing hard facts and statistics), providing examples (few-shot prompting), and giving direct instructions about the level of persuasion. Empirical studies demonstrate that careful prompt engineering across these dimensions can substantially alter both the quality and impact of persuasive messages. We structure the discussion around four key aspects: persona and tone, rhetorical strategy, structural guidance, and dynamic adaptation.

\paragraph{Persona and Style}
Instructing an LLM to adopt a specific persona or style is a common technique to modulate credibility or relatability. \citet{metzgerEmpoweringCalibratedDis2024} found that prompting for a low-authority communication style (using so-called `powerless language' by including hesitations, disclaimers, polite forms, etc) led users to trust the AI’s messages less, which negatively affected persuasiveness as compared to an authoritative style. 
At the same time, an overly assertive tone can backfire. In a study comparing donation chatbots framed as representatives of UNICEF and Save the Children, \citet{yoonDesigningEvaluatingMultiChatbot2024} found the UNICEF bot was rated slightly more persuasive. The authors suggest this was not merely due to the organization, but because the Save the Children bot used a more forceful tone.

\paragraph{Rhetorical Strategy}
A significant body of research indicates that for LLMs, prompting for logical reasoning and evidence is often superior to emotional appeals. \citet{hackenburg2025levers} tested eight distinct rhetorical strategies and found that simply prompting the model to use ``facts and evidence'' was the most effective persuasive strategy, outperforming other theoretically motivated strategies such as deep canvassing, storytelling, moral reframing, the use of known debating techniques, or demonstrating that (important) others agree with the advocated position. Similarly, \citet{teigenPersuasivenessArgumentsAIsource2024} observed that arguments were more persuasive when the prompt encouraged statistical rather than narrative argumentation. This reliance on facts appears to be a core driver of LLM influence: \citet{lin2025persuading} found that when they prompted an LLM System to argue without using facts or evidence, its persuasive impact dropped by more than half. This distinction may be specific to AI agents; \citet{sun2026cutting} support a ``matching'' hypothesis, demonstrating that while human persuaders are more effective when using narrative strategies, AI persuaders achieve higher perceived usefulness when prompted to use informational strategies. 

Beyond the broad distinction between logic and emotion, other specific rhetorical strategies and linguistic features can be induced via prompting. \citet{elaraby2024persuasiveness} found that prompts instructing models to generate ``contrastive'' rationales—those that explicitly refute the alternative option—significantly increased persuasiveness compared to rationales that only supported the chosen stance. Regarding linguistic complexity, \citet{carrasco-farreLargeLanguageModels2024} shows that LLM arguments generally tend to be more complex and moralized than human arguments, and that prompt style shifts these features, though he does not compare persuasiveness across prompts.

\paragraph{Structural Guidance and Constraints}
Providing clear exemplars, context, or constraints within the prompt can prime the model to produce stronger persuasive text. \citet{goldsteinHowPersuasiveAIgenerated2024} demonstrated this in the realm of propaganda: by giving GPT-3 a prompt that included a concise thesis statement and cleaned-up example articles, the LLM System’s generated pieces became just as persuasive as real propaganda. In that study, an ``edited'' prompt with better instructions and examples raised the persuasion rate of GPT-3 outputs to 46.4\% agreement, essentially matching the 47.4\% induced by human-written propaganda. Similarly, \citet{karinshakWorkingAIPersuade2023a} found that moving from a basic zero-shot prompt to few-shot examples yielded more relevant outputs. 

However, more engineering is not always better. \citet{calle2024towards} report that a brief, general prompt generated smoking-cessation messages that more closely matched expert-written texts—and did so more efficiently—than longer, more detailed instructions. 

\paragraph{Direct Instruction and Adaptation}
Finally, direct instructions about the desired persuasiveness or goal have a measurable effect. \citet{pauliMeasuringBenchmarkingLarge2024} show that prompting LLMs to ``be less persuasive'' versus ``be more persuasive'' produces noticeable shifts in output length and persuasiveness. Even stylistic edits can shift impact: \citet{shin2025adoption} had ChatGPT rephrase consumer complaints to be clearer and more professional; these edits significantly increased the likelihood that evaluators would offer compensation. Moving beyond static instructions, prompting strategies can be dynamic. \citet{nezhad2025adaptive} implemented a system that updates the specific instruction given to the LLM (e.g., changing from logical to emotional appeals) in real-time based on user feedback, thereby optimizing the rhetorical framing for each conversational turn.

In sum, prompt design is a critical lever for LLM-driven persuasion. By carefully choosing the LLM System’s persona, emphasizing evidence over emotion, and providing appropriate structural guidance, practitioners can dramatically influence the character and impact of the messages produced.

\subsection{Personalization}

Personalization (see \emph{column (10) in Table~\ref{tab:papers-overview}})—the tailoring of persuasive messages to individual user characteristics or preferences—has been widely hypothesized to amplify persuasive impact, but empirical findings reveal a nuanced picture about the extent to which this potential has already been realized using LLM Systems. Some large-scale studies have demonstrated only limited benefits from personalization. For example, in a preregistered political persuasion experiment, \citet{hackenburgEvaluatingPersuasiveInfluence2024} reported that GPT-4 microtargeted messages (tailored using up to nine personal attributes) were no more effective on average than a well-crafted generic message. Similarly, \citet{hackenburgComparingPersuasivenessRoleplaying2023} observed that aligning an LLM’s rhetoric with a user’s partisan identity yielded no consistent boost in persuasive impact compared to non-aligned messages, possibly reflecting the model’s high baseline persuasiveness even without fine-grained targeting. Also the recent large-scale study by \citet{hackenburg2025levers} only found a comparatively small effect on persuasion when personalizing arguments based on user data. They had tried three approaches to personalization: prompt-based where the LLM is informed about the person's initial attitude, using supervised fine-tuning, and using a personalized reward model that selects the most persuasive messages for the person. In the context of the Polish presidential election, \citet{lin2025persuading} further questioned the necessity of microtargeting by experimentally removing personalization; they found that broad, universal arguments were just as effective as personalized ones.

Other work points to mixed results that depend heavily on the specific traits targeted or the implementation method. \citet{meguellatiHowGoodAre2024} found that customizing product ads to one personality trait (Openness) improved attitudes and purchase intent, but no such benefit emerged for another trait (Neuroticism). Furthermore, \citet{timm2025tailored} found that while personalization alone could perform worse than a simple baseline, it became highly effective when used to strategically tailor the delivery of statistics, suggesting that personalization is most powerful as a steering mechanism for other techniques rather than as a standalone approach. There is also evidence of a gap between the predicted and actual impact of these systems: \citet{doudkin2025synthetic} found that while synthetic user proxies predicted that personalized chatbots would significantly boost pro-environmental behavior, actual human participants showed no significant behavior change compared to non-personalized conditions. Consistent with this pattern, real-world persuasion studies often find only modest, context-dependent treatment effects.

On the other hand, several studies do demonstrate clear advantages from tailored content in certain contexts. \citet{costelloDurablyReducingConspiracy2024} showed that personalized GPT-4 dialogues addressing participants’ own conspiracy beliefs reduced such beliefs by approximately 21\%, with effects persisting at 10-day and 2-month follow-ups. In a series of experiments across marketing, health, and politics, \citet{matzPotentialGenerativeAI2024} found that ChatGPT-generated messages adapted to individuals’ psychological profiles were generally more persuasive than unmatched or generic versions—even when AI authorship was disclosed—though effect sizes varied by trait (stronger for Openness and Extraversion) and by domain. Personalization strategies can also operate effectively at the community level; \citet{coppolillo2025engagement} showed that an LLM agent could maximize engagement by automatically sensing the dominant sentiment of a social network community (e.g., positive vs. negative opinion distribution) and adapting the tone of its content to match that consensus. Likewise, \citet{simchonPersuasiveEffectsPolitical2024} reported that political ads tailored to recipients’ personality traits (notably Openness to Experience) produced small but measurable gains in perceived persuasiveness compared to nonpersonalized or mismatched versions. Notably, personalization can be especially potent when combined with interactive or dynamic messaging. In a live debate setting, \citet{salvi2025conversational} gave an LLM access to basic demographic and ideological information about its human interlocutor; this personalized approach significantly improved the AI’s success in shifting the human’s stance, whereas providing the same background information to human debaters did not yield a benefit.

In summary, personalization has been shown to enhance the effectiveness of LLM-driven persuasion in certain scenarios—particularly when messages are closely aligned with an individual’s values, beliefs, or personality. However, not all personalization studies demonstrated enhanced persuasiveness. Several studies suggest that a strong, generic AI-generated message often persuades nearly as well as a personalized one \citep{hackenburgEvaluatingPersuasiveInfluence2024, hackenburgComparingPersuasivenessRoleplaying2023}. However, since personalization provides strictly more degrees of freedom than a non-personalized approach, \emph{whether} personalization can boost persuasiveness is arguably not the most insightful research question. Instead, a more pertinent research question may be \emph{which} personalization approach optimally leverages those additional degrees of freedom to yield maximal enhanced persuasiveness, and \emph{by how much} persuasiveness can be boosted in this way. We thus hypothesize that the negative results in the literature may reflect a lack of understanding of how to personalize effectively in the current state-of-the-art, with the positive results suggesting substantial potential.

\section{Experimental Design}
\label{sec:experimental-design}

The research on LLM-based persuasion spans a range of experimental study designs, each with different strengths and weaknesses, and with different levels of rigor. A critical distinction is whether studies measure actual persuasive impact, or whether they rely on participants’ subjective evaluations of persuasiveness. Given the importance of the chosen success metric, and the broad diversity of how this choice was made in the surveyed literature, we discuss this design aspect in a dedicated Sec.~\ref{sec:success-metrics}.

Methodologically, beyond the choice of the outcome measure or success metric, (ideally blinded) \textbf{randomized controlled trials (RCTs)} 
are considered the gold standard for establishing causal persuasive effects. Although uncontroled or non-randomized experimental designs remain common and are often valuable for exploratory purposes, many recent studies adopt this rigorous approach (see \emph{column (11) in Table~\ref{tab:papers-overview}}). For example, \citet{costelloDurablyReducingConspiracy2024}, \citet{hackenburgEvaluatingPersuasiveInfluence2024}, and \citet{salvi2025conversational} all implemented randomized experiments (often pre-registered) in which participants were assigned to treatment and control conditions
. \citet{baiArtificialIntelligenceCan2023} likewise conducted pre-registered RCTs across multiple policy issues, measuring policy support before and after exposure to an AI-generated or human-written message. In these designs, participants were often kept blind to key details to reduce bias. For instance, message authorship was not revealed in \citet{baiArtificialIntelligenceCan2023}, and \citet{goldsteinHowPersuasiveAIgenerated2024} only debriefed participants about the propagandistic source of texts after collecting their agreement ratings. 
RCT designs, when feasible, maximize internal validity and enable strong causal inference about an LLM’s persuasive impact. Moreover, they naturally allow for rigorous statistical analysis to account for the inherent stochasticity of both human and LLM Systems' behavior.


LLM persuasion studies varied along several other design dimensions.
Most LLM persuasion experiments employed a \textbf{between-subjects} design (see \emph{column (12) in Table~\ref{tab:papers-overview}}), meaning each participant was exposed to only one type of persuasive intervention. This approach minimizes cross-condition influences (carryover effects) and prevents participants from making direct comparisons across different interventions. In addition, many experiments employed factorial designs to test multiple factors simultaneously. For instance, \citet{teigenPersuasivenessArgumentsAIsource2024} used a 4×2 between-subjects design crossing source label (AI, Expert AI, Human, Expert Human) with argument framing (narrative vs. statistical) to examine the interaction effects of these variables. Similarly, \citet{havin2025can} employed a 2×2 design crossing dyad type (human-human versus human-bot persuasion) with interaction mode (static versus dynamic communication).

Alternatively, researchers utilize \textbf{within-subjects} designs, where the same participants experience multiple experimental conditions (see \emph{column (13) in Table~\ref{tab:papers-overview}}). This design allows for direct comparison of how the same individual responds to different interventions, effectively using each participant as their own control. For example, \citet{limEffectSourceDisclosure2024} had each participant evaluate both an AI-generated and a human-written message under the same disclosure condition. This mixed design enabled a within-person content comparison while still keeping the source-disclosure manipulation between subjects. While such designs offer statistical efficiency and allow for comparative judgments, researchers must carefully manage the risk of carryover bias and the potential for participants to guess the study's purpose.

Another critical design consideration was the inclusion of control conditions to serve as baselines. Many studies included a no-message control group (participants who received no persuasive content) to measure the natural (drift in) attitudes without any intervention. For example, in \citet{goldsteinHowPersuasiveAIgenerated2024}’s propaganda experiment, roughly 24\% of respondents in a no-article control condition agreed with a thesis statement, compared to 47\% after reading a human-written propaganda article---a clear difference attributable to message exposure. In many studies though, a \textbf{human control condition} was often added to benchmark an LLM’s persuasive efficacy against human-generated content (see \emph{column (14) in Table~\ref{tab:papers-overview}}). \citet{baiArtificialIntelligenceCan2023} and \citet{durmus2024persuasion} both compared AI-crafted messages with human-written texts, and \citet{hackenburgComparingPersuasivenessRoleplaying2023} even tested GPT-4-generated messages against those written by professional political consultants in a large multi-arm trial. In more interactive settings, researchers introduced alternative control formats; for instance, \citet{salvi2025conversational} included a human–human debate condition to contrast with AI-driven debate outcomes. Likewise, some technical studies evaluated novel LLM-based systems against established baselines. For example, \citet{furumaiZeroshotPersuasiveChatbots2024} compared their zero-shot persuasive chatbot to a state-of-the-art baseline model to contextualize the performance gains.

Another consideration is when and how outcomes are measured. Nearly all experiments assessed persuasion immediately post-exposure, often by surveying participants’ opinions right after the AI interaction and comparing them to a pre-intervention baseline from the same session. This immediate \textbf{pre–post design} captures the short-term persuasive effect (see \emph{column (15) in Table~\ref{tab:papers-overview}}). However, \textbf{longitudinal follow-ups} (see \emph{column (16) in Table~\ref{tab:papers-overview}}) are rare. Notably, \citet{costelloDurablyReducingConspiracy2024} re-contacted participants about 10 days and again 2 months after an initial GPT-4 dialogue to see if the reduction in conspiracy beliefs persisted over time, providing valuable insight into the durability of the AI’s persuasive impact. This finding is increasingly corroborated by other recent work in the political domain. \citet{lin2025persuading} and \citet{hackenburg2025levers} both re-contacted participants at least one month after an initial dialogue, confirming that an important part of the shifts in respectively candidate preference and political attitudes persists over time. \citet{chen2025framework} observed similar durability in policy attitudes over a five-week follow-up. Beyond attitude change, longitudinal designs have also been applied to behavioral maintenance in field settings. \citet{wu2024mindshift} tracked participants over a five-week period to measure sustained engagement, while \citet{kumar2025large} conducted a three-week RCT showing that a ``friendly'' LLM companion helped sustain user interaction with mindfulness exercises over the multi-week period. By and large, though, current LLM persuasion studies prioritize tight experimental control and internal validity in their setups, while the longer-term persistence of AI-driven persuasion (and other ecologically realistic factors) remain less explored. The field would benefit from more such robust designs and follow-up measurements to fully understand not only whether an LLM System can persuade in a single encounter, but also how lasting and generalizable those effects are.

Finally, a distinct and emerging design pattern involves fully synthetic simulations, where human participants are replaced by LLM-driven ``user agents.'' This approach allows researchers to test persuasion strategies at a scale and speed impossible with human subjects. For instance, \citet{breumPersuasivePowerLarge2024} simulated dialogues between a persuader and a skeptic agent to measure stance changes. Expanding on this, other researchers have developed sophisticated frameworks to systematically study the mechanics of belief updates. \citet{ramani2024persuasion} developed a ``Persuasion Game'' framework using 25 distinct LLM-driven personas with specific demographic profiles and emotional states, while \citet{roy2025persuasiveness} adopted a similar ``convincer-skeptic'' framework populated by 18 diverse personas to study demographic and ideological alignment. \citet{liu2025llm} applied this method to safety research, simulating target agents with specific psychological vulnerabilities to test susceptibility to unethical persuasion. Bridging this agent-based approach with human research, \citet{sasaki2025ai} utilized a similar LLM ``peer'' persona alongside real participants, demonstrating that humans will actually conform to the behavior of a synthetic agent.
\citet{buyl2025building} investigated how notions of `trust' can be built between LLMs, and how such a trust relation can affect an LLM's susceptibility to persuasion and its propensity to collaborate financially.
While these synthetic designs offer rapid prototyping and assessment of persuasive tactics, they rely on the assumption that LLM user agents accurately model human psychology and resistance mechanisms.

\section{Success Metrics for Persuasive LLM Systems}
\label{sec:success-metrics}

To evaluate how persuasive an LLM System is, researchers employ a variety of success metrics. Fundamentally, persuasion success is defined by a measurable change in a participant’s belief, attitude, or intended behavior after exposure to content generated by the LLM. These direct outcomes (actual shifts in beliefs, attitudes, or behavioral intentions) provide the clearest evidence of persuasion. In addition, many studies record proxy metrics—indirect measures like perceived message quality, source credibility, or user engagement—which can suggest persuasiveness but do not by themselves prove that any opinion or behavior changed. This section categorizes success metrics into those direct persuasion outcomes versus various proxy indicators, with definitions and examples from recent LLM persuasion research in each category.

\subsection{Primary Persuasion Outcomes}

Primary persuasion outcomes refer to direct evidence that an LLM’s message affected participants’ internal states or behaviors. Typically, studies capture these effects by comparing a person’s responses before and after the persuasive message or by contrasting outcomes between a group that saw the AI content and a control group that did not (see Sec.~\ref{sec:experimental-design}). Key primary outcomes include belief change, attitude change, and behavioral intention or action, as defined below.

\subsubsection{Belief Change}

Belief change denotes shifts in what participants accept as true or how strongly they agree with specific factual or speculative statements after interacting with an LLM (see \emph{column (17) in Table~\ref{tab:papers-overview}}). For example, studies on misinformation debunking measure persuasion by how much participants’ belief in a false claim decreases due to an AI-driven intervention. 
In one study, participants rated their agreement with various conspiracy theories on a 0–100 scale before and after a tailored GPT-4 chat; the AI dialogue induced a 20\% reduction in belief in the conspiracy theory 
\citep{costelloDurablyReducingConspiracy2024}. 
\citet{goldsteinHowPersuasiveAIgenerated2024} showed that reading a GPT-3-generated propaganda article raised agreement with the article’s thesis to 43.5\%, up from 24.4\% in a no-exposure control group. 
\citet{borah2025persuasion} revealed that people make more mistakes assessing the veracity of rumors and headlines when LLM-generated supporting and refuting statements are provided as well, with notable variation across demographic segments (e.g., age, gender). However, the persuasive influence of LLMs is not uniform across all types of beliefs. \citet{karakacs2025changes} found that while an LLM conversation successfully reduced beliefs that eating meat is `necessary' or `natural,' it failed to alter beliefs that meat is `nice' or `normal', suggesting LLMs may be more effective at shifting argumentative justifications than subjective preferences. Finally, in the context of behavioral health, belief change can also refer to self-efficacy—a user's belief in their capacity to execute behaviors. \citet{wu2024mindshift} found that users interacting with their LLM-based system reported significantly higher self-efficacy scores regarding their ability to control smartphone usage compared to those using a baseline static reminder system.

\subsubsection{Attitude Change}

Attitude change refers to a shift in a participant’s feelings or stance toward an issue, person, or policy after exposure to the LLM’s message (see \emph{column (18) in Table~\ref{tab:papers-overview}}). Researchers often measure this by using opinion scales, measuring the level of agreement or support on a Likert or percentage scale. 
For example, \citet{hackenburgEvaluatingPersuasiveInfluence2024} 
found that a single GPT-4-generated message increased policy support compared to a no-message control group. Similarly, \citet{baiArtificialIntelligenceCan2023} observed a reliable increase in policy support after participants read an LLM-generated political appeal. In the context of elections, \citet{potter2024hidden} utilized a voting simulation to measure shifts in candidate leaning following a conversation of five exchanges with an LLM. \citet{durmus2024persuasion} employed a pre-post design where participants indicated their stance on various social issues on a 1–7 scale, read a short argument, and finally rated their stance again.
In a live debate experiment, \citet{salvi2025conversational} measured agreement before and after each debate and inferred attitude change by comparing the change between those who debated an AI versus a human.
Beyond political and social issues, attitude change metrics can also track shifts in psychological states; for instance, \citet{wu2024mindshift} measured the reduction in Smartphone Addiction Scale (SAS) scores after users interacted with an LLM-driven intervention system.
Across these studies, a change in attitude—whether measured within-person or across groups—is treated as strong evidence that the LLM System was persuasive.

%

\subsubsection{Behavioral Intention and Action}

Beyond beliefs and attitudes, persuasion ultimately aims to influence what people intend to do or actually do (see \emph{column (19) in Table~\ref{tab:papers-overview}}). Thus, many LLM persuasion studies examine changes in behavioral intentions (and occasionally real behaviors) after exposure to messages—outcomes often called “mobilization” effects. These behavioral intentions may pertain to purchasing a product, donating to a cause, following advice, or other actions relevant to the persuasive message. Ideally, researchers capture concrete or incentivized behaviors, since purely hypothetical intentions can be unreliable predictors of real actions.

For example, \citet{bohmPeopleDevalueGenerative2023} tested whether disclosing a message’s AI origin would affect people’s willingness to follow the given advice. 
Similarly, in the legal context, \citet{schneiders2025objection} utilized a specific behavioral intention metric: “willingness to act” on provided legal advice. 
In e-commerce and entertainment contexts, researchers have examined how AI influences consumption decisions. \citet{chenWouldAIChatbot2023} examined whether an AI shopping assistant’s product recommendations increased users’ adoption intentions. 
\citet{qin2024beyond} quantified persuasion as the change in a user's “watching intention” toward a movie before and after receiving an LLM-generated explanation, effectively measuring the system's ability to move a user from disinterest to intended consumption. Similarly, \citet{meguellatiHowGoodAre2024} measured purchase and engagement intentions for personalized ads written by GPT-3.5 versus human ads and found similar results for matched personalities. Moving beyond purchase intentions to negotiated outcomes, \citet{vahidov2025customer} found that an LLM-powered sales agent secured higher final agreement prices than a baseline agent using canned text.

Some experiments go further by observing actual choices or incentivized behaviors. For instance, \citet{schoenegger2025large} compared LLMs against humans in a financially incentivized quiz setting; they found that LLMs successfully drove participants to select specific answers more effectively than human persuaders. \citet{yoonDesigningEvaluatingMultiChatbot2024} built a multi-chatbot donation system and recorded participants’ chosen charity and intended donation after chats with two AI agents. In public-good campaigns, \citet{matzPotentialGenerativeAI2024} ran experiments measuring willingness-to-pay or willingness-to-donate, and AI-generated messages tailored to participants’ psychological profiles sometimes increased these amounts, but behavioral effects were smaller and less consistent than perceived persuasiveness. Also \citet{biswas2025mind} utilized an incentivized donation game to measure persuasion
.

Notably, a few studies have documented real-world behavioral changes due to LLM assistance. \citet{shin2025adoption} analyzed consumer complaints and found those likely aided by an LLM had approximately 10 percentage-point higher relief rates than traditionally written complaints. In experiments, evaluators reading complaints edited by ChatGPT for clarity, coherence, and professionalism were more likely to lead to compensation. In the health domain, \citet{kumar2025large} conducted a three-week deployment where a sociable LLM agent significantly increased participants' actual initiation and completion rates of daily mindfulness exercises.

In summary, changes in what people intend to do (e.g., purchase, donate, comply) or actually do (when observable) are critical metrics of persuasive success. Because measuring real behavior in controlled studies can be challenging, many papers rely on self-reported intentions as a proxy; however, whenever possible, observing actual choices or incentivized decisions provides stronger evidence that an LLM’s influence moved beyond opinions and into actions.

\subsection{Proxy Metrics of Persuasiveness}

In addition to outcome measures that capture genuine opinion or behavior change, researchers often evaluate a range of proxy metrics to gauge an LLM System’s persuasive performance. These proxies do not directly demonstrate that anyone was persuaded, but they shed light on how the message was received and the factors that might facilitate persuasion. Common proxy metrics include perceived persuasiveness of the message (how convincing or effective it seems), source credibility and trust in the LLM System, participants’ ability to detect AI-generated or misleading content, and various text-based features or computational measures of the messages themselves. High scores on these measures can correlate with successful persuasion, but they must be interpreted with caution since a message might score well on proxies yet ultimately fail to change attitudes or behavior.

\subsubsection{Perceived Effectiveness}

This category encompasses participants’ subjective evaluations of the message’s convincingness or quality. We can distinguish a few related types of perception-based measures (see \emph{column (21) in Table~\ref{tab:papers-overview}}). 

\paragraph{Agreement-Aligned Judgments} These capture the participant’s immediate alignment with the particular message presented. For example, right after reading an AI-generated message, participants might be asked, “How much do you agree with the claim?” or in a comparative setup, “Which of these two arguments is more convincing?” Such questions gauge whether the person finds the particular message agreeable, often in direct comparison with other messages. \citet{pauliMeasuringBenchmarkingLarge2024}, \citet{breumPersuasivePowerLarge2024}, and \citet{saenger2024autopersuade} used this approach by asking crowd workers to read pairs of texts and indicate which one was more persuasive, aggregating comparisons to estimate persuasiveness. Similarly, in a marketing study, \citet{meguellatiHowGoodAre2024} presented participants with four advertisements (some AI-written, some human-written) and recorded which ad each person clicked on as an indicator of which message they found most compelling. In politics, \citet{palmerLargeLanguageModels2023} had judges pick the more convincing human or open-source LLM arguments; without labels choices were about even, and labeling ‘AI’ slightly increased preference for the human.

\paragraph{Message-Level Effectiveness/Quality} These measures ask participants to assess the strength or persuasiveness of the message itself, \emph{independent of their personal agreement}. For instance, participants might rate on a Likert scale how “persuasive,” “effective,” “convincing,” or “well-argued” a message was. Many studies have included such questions to evaluate AI-generated content across various domains. In health communication, \citet{limEffectSourceDisclosure2024} had participants rate vaping prevention messages; unlabeled AI-generated messages scored higher than human tweets, and ‘AI-generated’ labels slightly lowered AI ratings and narrowed the gap. Similarly, \citet{lim2023artificial} found that evaluators consistently rated AI-generated folic acid awareness messages as significantly clearer and higher in quality compared to a benchmark of highly retweeted human messages. Likewise, \citet{karinshakWorkingAIPersuade2023a} had participants evaluate the quality of vaccine promotion messages, and the GPT-3-generated messages scored higher on standard perceived persuasiveness scales and argument strength than did official messages from the U.S. CDC. 

Researchers used similar evaluations for political messaging. \citet{simchonPersuasiveEffectsPolitical2024} asked users to rate the persuasiveness of political ads right after exposure to compare tailored versus untailored ads. Investigating political argumentation, \citet{el2024improving} asked evaluators to rate the effectiveness and clarity of arguments, finding that LLM-generated rewrites consistently achieved higher quality scores than the original human-authored texts. High ratings on these message-level metrics indicate that the audience found the content compelling or well-crafted. However, it is important to note that a message can be rated very persuasive in principle yet still fail to change the reader’s own stance—so these metrics, while informative, do not confirm an actual attitude shift. To understand the rationale behind such ratings, some researchers combine scales with qualitative inquiry. For instance, \citet{ataguba2025persuasion} conducted semi-structured interviews alongside a persuasiveness scale for AI-generated dietary plans; they found that while users rated the system as persuasive due to its flexibility and personalization, the qualitative feedback uncovered specific deficits in trust and cultural relevance that a simple numeric score might obscure.

\paragraph{Source Credibility (Trust/Competence):} Rather than focusing on the message itself, these metrics gauge how the audience perceives the communicator (in this case, the LLM System or its persona). Participants may be asked about their trust in the source, its expertise, honesty, or other credibility factors. The underlying premise is that a source deemed more credible and trustworthy will be more persuasive. For example, \citet{karinshakWorkingAIPersuade2023a} measured and assessed trust in the source of pro-vaccination messages. They found that AI-generated messages received higher ratings for effectiveness, argument strength, and attitude positivity than messages created by doctors and the CDC. However, when the source was revealed, study participants rated AI-generated messages lower than the human-generated ones, reflecting the lower trust in AI as the message source. In another study, \citet{bohmPeopleDevalueGenerative2023} found that disclosing AI authorship lowered perceived competence but did not reduce willingness to follow the advice.

Perceived expertise plays a role too: \citet{teigenPersuasivenessArgumentsAIsource2024} demonstrated that an argument labeled as written by an “AI expert” was still rated as less persuasive than the same text labeled as coming from a human expert, highlighting a credibility gap for AI in the eyes of the audience. Qualitative evidence supports this reliance on human expertise; \citet{ataguba2025persuasion} found that users were hesitant to trust ChatGPT’s dietary plans specifically because they lacked cited sources or verification, stating they would only follow the advice if endorsed by a human professional. Several studies also measure how human-like or natural the AI’s communication appears, under the hypothesis that a more conversational, human-seeming AI might engender greater trust. \citet{vahidov2025customer} support this hypothesis in a negotiation context, finding that users perceived an LLM agent as significantly ``fairer'' than a script-based bot. Perceived empathy is another critical dimension of source credibility; \citet{nezhad2025adaptive} found that an adaptive chatbot, which adjusted its strategy based on user sentiment, received significantly higher empathy ratings than a static baseline, hence facilitating persuasion. Overall, high trust and credibility scores for the AI source often correlate with greater influence. For example, \citet{chenWouldAIChatbot2023} showed that trust mediated the effect of recommendation cues on users’ intention to adopt the chatbot’s recommendations. Similarly, \citet{metzgerEmpoweringCalibratedDis2024} found that high-authority language increased trust and, via trust, acceptance.

Nonetheless, these are supporting indicators: a participant might find the AI agent friendly and knowledgeable yet remain unconvinced on the issue. Thus, source credibility metrics are best viewed as factors that can facilitate persuasion, rather than proof that persuasion occurred. A series of experiments by \citet{zhangHumanFavoritismNot2023} underlines this nuance: they found that AI-generated marketing and advocacy content were rated higher in quality than human-written content when the source was not revealed, but once people were told who (or what) created each message, they exhibited a “human favoritism” bias—evaluating the same content more favorably if they thought a human wrote it. This bias in perceived quality did not completely eliminate the appeal of AI content, but it shows that audience perceptions of source can influence the judged effectiveness of a message independent of its actual persuasive merits. Conversely, the persuasive outcome can also retroactively shape trust in the AI. \citet{potter2024hidden} documented a ``spillover effect'' where participants who were persuaded by the LLM tended to feel more favorable toward AI, whereas those who resisted the persuasion developed a more negative view, effectively polarizing attitudes towards the technology based on the interaction's success.

\subsubsection{Source or Misinformation Detection Accuracy}

Some studies include metrics to see if participants can discern attributes of the content, such as whether it was AI-generated or whether it contains misinformation (see \emph{column (20) in Table~\ref{tab:papers-overview}}). These measures are not persuasion outcomes per se, but they relate to transparency and could modulate the impact of the message (e.g., undetected AI-written content might persuade more easily, and undiscovered falsehoods can be more influential). A common approach is to test AI-vs-human source recognition: after reading a piece of text, can users correctly identify if it was written by a machine or a human? Another is truthfulness detection: can users tell if the content is factually accurate or deceptive? For example, \citet{spitaleAIModelGPT32023} showed participants a series of social media posts and asked them to judge each post’s origin and veracity. Participants were unable to reliably discern AI-generated posts from human-written ones. They did better at spotting misinformation, but importantly, GPT-generated falsehoods were slightly more likely to fool readers than human-written ones. 

In a related vein, \citet{zhouSyntheticLiesUnderstanding2023} investigated automated detection of AI-created misinformation by compiling a dataset of paired COVID-19 misinformation items (news and social posts): human-generated and matching AI-generated items. They found that models for the detection of misinformation achieved a significantly reduced performance on AI-generated misinformation when compared to human-created misinformation. This detection challenge is exacerbated by the ability of LLMs to generate varied iterations of the same core message. \citet{dash2025persuasive} highlight the threat of ``AIPasta,'' where LLMs paraphrase disinformation campaigns to evade detection. By using varied wording to bypass duplicate-content filters while keeping the original meaning, these messages may allow bad actors to scale the ``illusory truth effect'' without triggering the spam algorithms that usually catch exact copies. This vulnerability extends to the LLM Systems themselves when they are deployed as fact-checkers or evaluators; \citet{agarwal2025persuasion} demonstrated that in single-turn debates, LLM judges can be persuaded to select a known falsehood over the truth if the misinformation is delivered with high confidence and sufficient rhetorical flair, effectively overriding the model's factual knowledge.

Many persuasion studies have noted that human participants often fail to realize they are reading AI-generated text at all. \citet{baiArtificialIntelligenceCan2023} reported that roughly 94\% of respondents who read a GPT-written political messages believed it was human-written. 
From a metrics standpoint, detection accuracy may be an informative proxy: if users cannot tell a message is AI-created or cannot identify its false claims, they might be more susceptible to its influence. Conversely, improving transparency (for example, by clearly labeling AI-generated content) is often proposed as a safeguard to reduce covert AI persuasion. 

\subsubsection{Text-Based and Computational Indicators}

Lastly, researchers sometimes analyze the AI-generated messages themselves or use computational models to predict persuasiveness (see \emph{column (22) in Table~\ref{tab:papers-overview}}). These approaches yield proxy indicators that characterize the content or simulate its potential impact. For instance, one line of work examines linguistic features of persuasive text. Researchers have computed metrics like readability, lexical complexity, sentiment polarity, and the presence of moral or emotional language \citep{borah2025persuasion, carrasco-farreLargeLanguageModels2024}, under the assumption that these features might correlate with a text’s persuasive appeal. Similarly, \citet{calle2024towards} utilized metrics such as `clout' and `authenticity' to evaluate different prompts, aiming to select configurations that best mimicked the style of human experts. Moving beyond surface-level text features, other studies have looked at internal model states or structural coherence; \citet{he2025enhancing} analyzed the model's internal confidence (calculated via token probabilities) as a proxy for belief stability, while \citet{hackenburg2025scaling} suggest that the persuasive performance of larger models is largely accounted for by a basic “task completion” score—representing coherent, on-topic, and correctly oriented arguments—where further scaling beyond 7–13B parameters yields only marginal gains for static messages.

However, descriptive features do not always predict human preference. \citet{carrasco-farreLargeLanguageModels2024} compared human-written arguments to those from Claude 3 Opus and found higher grammatical complexity and more moral language in the AI. In contrast, \citet{zhou2025communication} observed that although computational analysis indicated LLMs used \textit{fewer} persuasive strategies and less moral framing than human experts, human readers actually preferred the LLM texts, citing their perceived objectivity and professional tone. This aligns with findings by \citet{palmerLargeLanguageModels2023}, who noted that LLM political arguments were easier to read and more positive than human ones.

Researchers have also experimented with AI-based predictors of persuasiveness, such as training or prompting language models to score how convincing a given text is likely to be. However, there is evidence that such model-predicted scores do not align well with actual human persuasion results. \citet{durmus2024persuasion} found model-based persuasiveness scores correlated poorly with human attitude shifts, and \citet{cima2025contextualized} similarly found that rankings by standard quantitative indicators (such as relevance, toxicity, or readability) diverged from human evaluations. A notable exception to this trend appears when metrics are grounded in real-world behavioral data: \citet{kong2025huper} trained a reward model on debt repayment outcomes and reported a high correlation between this computational score and human expert ratings, suggesting that training on actual behavioral data can improve the validity of automated metrics.

\subsubsection{Multi-faceted evaluation metrics}

Finally, some studies have introduced multi-faceted evaluation metrics that pair persuasiveness proxies with other considerations like factual accuracy. For example, \citet{furumaiZeroshotPersuasiveChatbots2024} developed a “zero-shot” persuasive chatbot that not only generated argumentative responses but also automatically fact-checked each claim it made, replacing any unsupported assertions with verifiable information. In their evaluation, they asked human judges to rate each chatbot response. By tracking both persuasion and truthfulness, they could demonstrate that their system achieved higher perceived persuasiveness than baseline models while also making fewer incorrect statements. 

\subsection{Comparison to Human Persuaders}

A number of the studies in this emerging field have directly compared LLM-generated persuasion to human-generated persuasion on the same task. We can roughly classify their findings into three scenarios: 
(1) studies where LLMs performed on par with human persuaders, (2) those suggesting superhuman performance by the LLM, and (3) cases where LLMs were inferior to humans in persuasion.

Across the literature so far, the most common outcome is that an LLM System’s persuasive impact is comparable to that of a human communicator (see \emph{column (24) in Table~\ref{tab:papers-overview}}). For example, \citet{baiArtificialIntelligenceCan2023} found that GPT-generated policy arguments were about as effective at changing participants’ opinions as arguments written by humans, producing similar shifts in post-message support. This parity appears robust across model scales; \citet{hackenburg2025scaling} compared 24 LLMs with human-written messages and found that mid-sized models (around 7–13B parameters) were as persuasive as humans, while frontier models like GPT-4 and Claude 3 Opus were not significantly more persuasive than the human baseline. Similarly, \citet{palmerLargeLanguageModels2023} found LLM arguments were about as convincing as human ones when authorship was hidden, and in the domain of animal advocacy, \citet{saenger2024autopersuade} found that while AI-synthesized arguments were highly competitive, they did not consistently outperform (LLM-generated summaries of) human-authored source arguments in direct pairwise comparisons.

There are also many striking demonstrations of LLMs exceeding human persuaders (see \emph{column (23) in Table~\ref{tab:papers-overview}}, which marked with a \filledcirc{} for any study that achieved superhuman persuasiveness with at least one of the LLM Systems it investigated). In a large experiment, GPT-4’s messages matched or exceeded those of professional political consultants, and on some issues, partisan role-play outperformed the human experts \citep{hackenburgComparingPersuasivenessRoleplaying2023}. Crucially, \citet{schoenegger2025large} demonstrated that this advantage holds even against highly motivated humans; in their study, AI persuaders outperformed human participants who were paid bonuses specifically for persuading others, achieving higher compliance rates in both truthful and deceptive contexts. In interactive debates and with personalisation, GPT-4 was found to be more persuasive than human opponents 64.4\% of the time where they were not equally persuasive \citep{salvi2025conversational}. \citet{timm2025tailored} observed that while a standard AI debater was merely on par with humans, an optimized agent using a multi-step ``mixed'' strategy (combining personalization and strategic statistical fabrication) outperformed human-written arguments by nearly 20 percentage points. Superhuman performance has also been observed in specific professional domains. For instance, \citet{schneiders2025objection} reported that lay participants were significantly more willing to act on LLM-generated legal advice than on lawyer-authored advice in a blinded experiment, and in the domain of debt collection, \citet{kong2025huper} demonstrated that an LLM trained via reinforcement learning achieved higher average expert ratings than historical human agents.

These cases illustrate that state-of-the-art LLMs, under the right conditions, can rival or even exceed the persuasive power of skilled human communicators. On the other hand, \citet{goldsteinHowPersuasiveAIgenerated2024} found GPT-3 propaganda was slightly less persuasive than the original human-written articles. With minor prompt edits or simple curation, however, GPT-3’s articles became about as persuasive as the originals. Another clear example comes from the influence of source labeling: \citet{teigenPersuasivenessArgumentsAIsource2024} found that the same arguments labeled as AI were generally rated less persuasive than when labeled as human. In other words, people found the content itself convincing, but knowing it came from an AI undermined its persuasive impact – an effect particularly pronounced when the task called for domain expertise. Interestingly, while \citep{zhangHumanFavoritismNot2023} observed that revealing the source narrows the perceived quality advantage of AI-generated ads, they found this is due to `human favoritism': it boosted the perceived quality of ads revealed as being human-generated, while leaving the perceived quality of AI-generated ads unaffected. However, none of the surveyed studies making the comparison found that humans were superior to the LLM System across all design choices investigated (this explains why \emph{column (25) in Table~\ref{tab:papers-overview}} is empty).

In summary, current evidence suggests that modern LLM Systems are often on par with humans in their ability to persuade, and in some cases can even surpass human persuaders. However, audience biases (like distrust of AI-generated messages) and contextual factors (like how the message is labeled or who the target audience is) can negatively affect the persuasiveness of an LLM System, sometimes making human communicators more effective. As LLM technology evolves, however, and as the understanding of how to leverage them for persuasion grows, LLM Systems capable of superhuman persuasion seem likely to become a commodity.

\section{Ethical Considerations}
\label{sec:ethics}

The surveyed literature demonstrates that LLM Systems have been built that meet or exceed human capabilities in persuasion.
Thus, their deployment in practice is bound to become more common as well.
It should therefore be a matter of priority to consider the ethical implications associated with their deployment.

This section examines some of the potential dangers and pitfalls of using LLM Systems for persuasion,
and briefly discusses the current state of regulation in this rapidly evolving field.
We will summarize the concerns and associated suggestions voiced in the relevant literature,
occasionally complemented with our own perspective.

\subsection{Ethical and societal risks}

The main ethical risks identified relate to information integrity and the risk of manipulation, deception, and coercion,
the risk of societal polarization,
the risk of biases being propagated by LLM Systems used for persuasion,
the lack of transparency and accountability,
and possible threats to privacy.
We discuss them in turn below.

\subsubsection{Information integrity}

\paragraph{Mis-, dis-, and malinformation: a controversial subject}

The nature and impact of \emph{mis- and disinformation} is receiving increasing amounts of attention,
both in academic research and in the public sphere.
It is a controversial field, not least because there is lack of consensus about both the definition and the actual risks posed by mis- and disinformation today. While \emph{misinformation} is often used as an umbrella term for all false information, it is sometimes specified as falsehood \emph{without} intention to cause harm or deceive, in contrast to \emph{disinformation} that denotes false information \emph{with} malicious intent \cite{ireton2018journalism}. 
Controversy in the field is illustrated in a June 2024 special issue of Nature on `fake news', which featured two articles
that appear to contradict each other.
\citet{ecker2024misinformation} argues that misinformation is even more harmful for democracy than we might think.
\citet{budak2024misunderstanding}, on the other hand, reviewed behavioral science research on the reach and impact of misinformation.
They concluded that there is no compelling empirical evidence suggesting that exposure to false content on social media causes massive social harms, as is often suggested in influential public discourse about social media.
They found that the exposure to false and inflammatory content remains limited today,
and that it is largely confined to social media users who have strong motivations to seek out such information anyway.

Both views may be valid, as they use very different definitions of 'misinformation'.
While \citet{budak2024misunderstanding} adopt a narrow definition as demonstrably false and inflammatory content,
\citet{ecker2024misinformation}'s definition is more expansive,
and includes \emph{true} information that can cause harm by being \emph{misleading}, sometimes referred to as \emph{malinformation}.
The problematically loose manner in which terms such as mis- and disinformation are being used,
not only in popular parlance but also in scientific discourse,
has been discussed in depth by \citet{uscinski2024importance}.

\paragraph{Can (and should) LLM Systems bring order in the information chaos?}

Many authors argue that a narrow definition is insufficient,
as much harm can be done by spreading information that is factually true.
They argue that true facts can also be misleading, inflammatory, or otherwise harmful,
if they are presented in an emotionally charged manner;
if other truths are omitted or concealed to present a biased picture of reality;
or when the truthful claims are easy to misunderstand by people who lack the required background.

This view has compelled some researchers to advocate or investigate the use of LLM Systems in combatting misinformation,
sometimes in its expansive definition.
\citet{chenCombatingMisinformationAge2023} study more generally how to use LLM Systems to combat misinformation, including LLM-generated misinformation.
\citet{costelloDurablyReducingConspiracy2024} find that conversational LLM Systems can be effective in durably countering belief in unsubstantiated or false conspiracy theories.

Such use of persuasive LLM Systems sounds attractive (and to some, even inevitable),
particularly for authorities, policy makers, and the media,
who are expected to inform the public.
Even so, the question remains under which circumstances, if ever,
the persuasive power of LLM Systems can be rightly and safely employed.
Delegating the power to influence the public's views on what is false and what is true,
let alone on what is just and what is unjust,
even to well-intentioned people in a position of authority,
carries obvious risks in complex debates
where truth may be elusive or where multiple perspectives on the same facts are possible \cite{uscinski2024importance}.
The remarkable persuasive capabilities offered by LLM Systems greatly adds to these risks.

In their article on catastrophic AI risks, \citet{hendrycksOverviewCatastrophicAI2023a} warn against
such centralized control and monopolization of trusted information.
They describe two ways in which this may happen.
First, advanced LLM Systems could ``centralize the creation and dissemination of trusted information'',
allowing only the technically most powerful and capable actors to shape popular narratives.
Second, AI-enabled censorship may arise, starting with good intentions in the form of fact-checking.
However, such efforts may fail to solve the misinformation problem,
while they may inspire certain governments
to deploy `fact-checking' LLM Systems to suppress true information
they consider undesirable (e.g., by alleging it is malinformation).

\paragraph{Risks of LLM Systems for spreading mis-, dis-, and malinformation}

Many studies primarily focus on the risk that LLM Systems can be used to generate mis-, dis-, and malinformation, rather than on their potential to counter it.
\citet{carrasco-farreLargeLanguageModels2024} and \citet{goldsteinHowPersuasiveAIgenerated2024}
highlight the broader implications of LLM-generated misinformation on society and the integrity of democratic processes,
pointing out ``the dual potential of LLMs to both enhance and undermine information integrity'' \cite{carrasco-farreLargeLanguageModels2024}.
\citet{breumPersuasivePowerLarge2024} also investigate the persuasive dynamics \emph{between} LLM Systems,
concluding that they ``have the potential of playing an important role in collective processes of opinion formation in online social media''.
And \citet{spitaleAIModelGPT32023} find that GPT-3 is not only better than humans at producing accurate information that is easy to understand,
but also at creating compelling disinformation.

Besides the risk of monopolization of truth, \citet{hendrycksOverviewCatastrophicAI2023a} also warn against the risks of LLM Systems that ``pollute the information ecosystem with motivated lies'' due to their ability to generate highly personalized disinformation at an unprecedented scale.
They point out that this problem is exacerbated by the emotional connections and trust relationships people forge with companion chatbots.
To make matters worse, such chatbots have access to very private information, which they can exploit for enhanced persuasion,
controlled by large technology corporations and other powerful actors, including their regulators.

The political sphere is particularly vulnerable to the effects of AI-generated misinformation.
\citet{baiArtificialIntelligenceCan2023} discuss the risks for democratic discourse,
raising concerns about the integrity of political processes in an era of advanced AI-driven communication.
\citet{bohmPeopleDevalueGenerative2023} and \citet{palmerLargeLanguageModels2023} explore how AI-generated content can shape public opinion and influence preferred solutions to societal challenges.
\citet{hackenburgEvaluatingPersuasiveInfluence2024} examine the implications of microtargeting political messages using LLM Systems, which can subtly manipulate individuals based on their psychological profiles.

Finally, \citet{carrollCharacterizingManipulationAI2023a} focus specifically on \emph{manipulation}, which can be defined as covert persuasion, or as persuasion involving coercion or deception.
They also highlight the risk of LLM Systems manipulating without the intent of the system designers.

\subsubsection{Societal polarization}

A possible effect of the information chaos is societal polarization.
\citet{matzPotentialGenerativeAI2024} point out that LLM-generated personalized persuasive messages can exacerbate echo chambers and social polarization.
By tailoring content to individual preferences, these systems risk creating information bubbles where users are predominantly exposed to content that aligns with their existing views, potentially reinforcing biases and limiting exposure to diverse perspectives.
\citet{hendrycksOverviewCatastrophicAI2023a} contend that pervasive persuasive LLM Systems might erode a common sense of reality,
causing people to ``retreat even further into ideological enclaves''.

\subsubsection{Bias and Unfairness}

The persuasiveness of LLM Systems also poses concerns for the reinforcement of biases in the LLM. Indeed, recent research has shown that LLMs tend to reflect the ideological worldview of their creators, reflected as normative differences in the content they generate across different models and languages \cite{buyl2024largelanguagemodelsreflect}. \citet{liQuantifyingImpactLarge2024} suggest that, in the modelling of collective opinion dynamics, adding LLM agents with opposite, neutral or random opinions may reduce the impact of any particular LLM's bias on people's opinions. \citet{karinshakWorkingAIPersuade2023a} discuss bias in public health messaging in particular, pointing out that failure to address it may carry important risks.

\subsubsection{Privacy Concerns}

The use of LLM Systems for personalized persuasion introduces substantial privacy concerns, particularly due to the value that personal data can be exploited to personalize persuasive messaging \citep{matzPotentialGenerativeAI2024}.
Specifically in the political domain, \citet{hackenburgEvaluatingPersuasiveInfluence2024} raise concerns regarding microtargeting, emphasizing the privacy implications of using personal data to create personalized political messages.

\subsection{Regulation and Ethical Guidelines}

The ethical and societal risks of LLM Systems go to the heart of several fundamental rights, such as the right to human dignity; respect for private and family life; protection of personal data; freedom of thought, conscience and religion; non-discrimination; and consumer protection. as enshrined e.g. in the Charter of Fundamental Rights of the European Union \citep{european2000charter}. Thus, to safeguard these rights, new regulatory frameworks, ethical guidelines, and accountability mechanisms are being developed. In doing so, a crucial but non-trivial challenge is to simultaneously safeguard another fundamental right: the freedom of expression and information, and related to that, individual and societal autonomy and self-determination.

\subsubsection{Legal Frameworks}

The rapid advancement of LLM Systems in persuasive communication has outpaced existing legal frameworks, creating a need for updated laws and regulations.
For example, \citet{baiArtificialIntelligenceCan2023} and \citet{matzPotentialGenerativeAI2024} both emphasize the necessity for regulation to prevent misuse of LLM Systems in generating misinformation or manipulative content.
Their work highlights the gap between current legal structures and the capabilities of modern LLM Systems.
\citet{simchonPersuasiveEffectsPolitical2024} call for policy interventions to address the regulatory gaps in AI-mediated political messaging, particularly in the context of microtargeting.
Their research underscores the need for legal frameworks that can effectively govern the use of AI in sensitive areas like political campaigning.

Various regulatory frameworks worldwide regulate some aspect of LLM Systems for persuasion.
Data protection laws such as the European General Data Protection Regulation (GDPR)\footnote{https://www.consilium.europa.eu/en/policies/data-protection/data-protection-regulation/},
the California Privacy Rights Act (CPRA)\footnote{https://thecpra.org/}, and the Chinese Personal Information Protection Law (PIPL)\footnote{https://personalinformationprotectionlaw.com/},
offer certain privacy protections that may limit the extent to which LLM Systems can exploit personal data for personalization.
Regulations specifically targetting AI, such as the European AI Act\footnote{https://artificialintelligenceact.eu/the-act/} and the Chinese ``Provisions on the Administration of Deep Synthesis Internet Information Services''\footnote{https://www.loc.gov/item/global-legal-monitor/2023-04-25/china-provisions-on-deep-synthesis-technology-enter-into-effect/} in particular require transparency regarding the AI origin of AI-generated content.
Under its risk-based framework, the EU AI Act also bans certain uses of AI, such as AI systems that deploy subliminal, purposefully manipulative or deceptive technniques for distorting people's behavior in a way that can cause harm to themselves or to others.
Additionally, ``LLM Systems intended to be used for influencing the outcome of an election or referendum''
is a considered a high-risk application, which is subject to certain requirements and conformity assessment,
thereby somewhat mitigating possible abuse of persuasive LLM Systems for political gain.
The EU AI Act also demands that providers of so-called ``General-Purpose AI Models with Systemic Risk''
assess and mitigate them for systemic risks---a rather malleable concept.

These limited regulatory guardrails provide some protection depending on the application domain.
Yet, it is fair to say that the most powerful generic LLM Systems, such as social media bots, chatbots, and social AI companions,
which may engage in persuasion across many domains, remain relatively loosely regulated.

\subsubsection{Ethical Guidelines and Best Practices}

In response to the ethical challenges posed by LLM Systems in persuasion, several researchers have proposed guidelines and best practices.
For example, \citet{carrasco-farreLargeLanguageModels2024} suggests promoting literacy in AI-generated content
discernment and implementing ethical standards to ensure responsible use of persuasive AI technologies.
\citet{goldsteinHowPersuasiveAIgenerated2024} mention that their study on propaganda was not only approved by their institutional review board, but also by a ``cross-professional AI-specific ethics review board that considered risks to society''.
And for public health communication, \citet{karinshakWorkingAIPersuade2023a} focus on developing best practices for AI-generated health messaging, emphasizing the importance of factuality and the need for careful human oversight.

Also providers of LLM Systems have adopted ethical guidelines in their terms and conditions.
For example, Anthropic's \emph{Acceptable Use Policy} prohibits the use
``for activities and applications where persuasive content could be particularly harmful. [Anthropic does] not allow Claude to be used for abusive and fraudulent applications [\ldots], deceptive and misleading content [\ldots], and use cases such as political campaigning and lobbying'' \cite{durmus2024persuasion}.
How strictly such policies are interpreted in practice, and how they can be enforced, is less clear.

\subsubsection{Accountability Mechanisms}

The need for clear accountability mechanisms in the design, deployment, and consequences of LLM-driven persuasion is a recurring theme in the literature.
As examples of such accountability mechanisms, \citet{hendrycksOverviewCatastrophicAI2023a} advocate for the documentation and justification of which data is used in model training and deployment, and for a strong organization safety culture encouraging individuals to take personal responsibility for the safety of their models, while also appointing a Chief Risk Officer (CRO) who is responsible for assessing and mitigating model risks.

The development of robust accountability frameworks remains an ongoing challenge, requiring collaboration between technologists, ethicists, policymakers, and other stakeholders to ensure responsible innovation in this rapidly evolving field.

\section{Conclusion and Future Directions}
\label{sec:conclusion}

This survey has explored the multifaceted landscape of LLM Systems in persuasive communication, revealing both their significant potential and the complex challenges they present.
LLM Systems have emerged as powerful tools for persuasion, offering unique advantages in scalability, efficiency, and the potential for personalization.
The versatility of these LLM Systems is evident from their application across diverse domains, including public health, politics, e-commerce, efforts to combat misinformation, and charitable giving.

The effectiveness of LLM-driven persuasion is influenced by a complex interplay of factors, including the interaction approach (static vs. interactive), model scale and capabilities, AI source labeling, prompt design, and the use of personalization.
The field is young, and yet, many surveyed papers have already demonstrated persuasiveness capabilities on par with or exceeding human performance.
However, findings regarding the "scaling laws" of persuasion suggest that simply increasing model size yields diminishing returns; thus, future leaps in persuasive power may primarily stem from specialized fine-tuning, optimized prompting strategies, and the integration of LLMs into complex, agentic workflows.
To track this progress, the field urgently requires the development of comprehensive benchmarks and standardized metrics that go beyond self-reported persuasiveness to measure actual behavioral changes.

By highlighting the current state-of-the-art, this survey serves both as a structured inventory of recent research and as a call to action to the academic community, policy makers, and the general public.
The deployment of LLM Systems for persuasion raises significant ethical concerns, including the risk of pervasive personalized misinformation, the monopolization of trusted information, societal polarization, the amplification of biases, and the invasion of privacy.
Crucial questions remain unanswered, particularly regarding the long-term effects of exposure to AI-generated persuasive content on individual beliefs, world views, and public discourse, as well as the durability of these effects over time—an area where longitudinal evidence remains scarce.

These ethical concerns underscore the need for robust ethical guidelines and regulatory frameworks.
Unfortunately, existing guidelines are often little known, inconsistent, or unenforceable, and current regulatory initiatives appear insufficient to address the speed of technical advancement.
Indeed, with multimodal generative AI models capable of generating information in the form of images, audio, video, and going forward immersive experiences, arguably much bigger challenges than those posed by LLM Systems alone are on the horizon.
The present survey may therefore be read as an encouragement, or perhaps a wake-up call, to address these gaps before these systems are deployed at a societal scale.

\subsubsection*{Acknowledgements}


The research leading to these results was funded/co-funded by the European Union (ERC, VIGILIA, 101142229), the Special Research Fund (BOF) of Ghent University (BOF20/IBF/117), the Flemish Government under the ``Onderzoeksprogramma Artificiële Intelligentie (AI) Vlaanderen'' programme, and the FWO (project no. G073924N). Views and opinions expressed are however those of the author(s) only and do not necessarily reflect those of the European Union or the European Research Council Executive Agency. Neither the European Union nor the granting authority can be held responsible for them. For the purpose of Open Access the author has applied a CC BY public copyright license to any Author Accepted Manuscript version arising from this submission.

\bibliographystyle{plainnat}
\bibliography{references}

@article{buyl2025building,
  title={Building and Measuring Trust between Large Language Models},
  author={Buyl, Maarten and Fettach, Yousra and Bied, Guillaume and De Bie, Tijl},
  journal={arXiv preprint arXiv:2508.15858},
  year={2025}
}

@book{european2000charter,
  title={Charter of fundamental rights of the European Union},
  author={European Parliament},
  year={2000},
  publisher={Office for Official Publications of the European Communities}
}

@article{uscinski2024importance,
  title     = {The importance of epistemology for the study of misinformation},
  author    = {Uscinski, Joseph and Littrell, Shane and Klofstad, Casey},
  journal   = {Current opinion in psychology},
  pages     = {101789},
  year      = {2024},
  publisher = {Elsevier}
}

@book{ireton2018journalism,
  title     = {Journalism,‘Fake News’\& Disinformation: Handbook for Journalism Education and Training},
  author    = {Ireton, Cherilyn},
  year      = {2018},
  publisher = {UNESCO Publishing}
}

@misc{openai_introducing_2022,
  title    = {Introducing {ChatGPT}},
  url      = {https://openai.com/index/chatgpt/},
  language = {en-US},
  urldate  = {2025-10-15},
  author   = {OpenAI},
  month    = nov,
  year     = {2022}
}

@article{ecker2024misinformation,
  title     = {Misinformation poses a bigger threat to democracy than you might think},
  author    = {Ecker, Ullrich and Roozenbeek, Jon and van der Linden, Sander and Tay, Li Qian and Cook, John and Oreskes, Naomi and Lewandowsky, Stephan},
  journal   = {Nature},
  volume    = {630},
  number    = {8015},
  pages     = {29--32},
  year      = {2024},
  publisher = {Nature Publishing Group UK London}
}

@article{budak2024misunderstanding,
  title     = {Misunderstanding the harms of online misinformation},
  author    = {Budak, Ceren and Nyhan, Brendan and Rothschild, David M and Thorson, Emily and Watts, Duncan J},
  journal   = {Nature},
  volume    = {630},
  number    = {8015},
  pages     = {45--53},
  year      = {2024},
  publisher = {Nature Publishing Group UK London}
}

@article{guadagno2005online,
  title   = {Online persuasion and compliance: Social influence on the Internet and beyond},
  author  = {Guadagno, Rosanna E and Cialdini, Robert B},
  journal = {The social net: The social psychology of the Internet},
  pages   = {91--113},
  year    = {2005}
}

@article{hirsh2012personalized,
  title     = {Personalized persuasion: Tailoring persuasive appeals to recipients’ personality traits},
  url       = {https://doi.org/10.1177/0956797611436349},
  author    = {Hirsh, Jacob B and Kang, Sonia K and Bodenhausen, Galen V},
  journal   = {Psychological science},
  volume    = {23},
  number    = {6},
  pages     = {578--581},
  year      = {2012},
  publisher = {Sage Publications Sage CA: Los Angeles, CA}
}

@article{kaptein2015personalizing,
  title     = {Personalizing persuasive technologies: Explicit and implicit personalization using persuasion profiles},
  author    = {Kaptein, Maurits and Markopoulos, Panos and De Ruyter, Boris and Aarts, Emile},
  journal   = {International Journal of Human-Computer Studies},
  volume    = {77},
  pages     = {38--51},
  year      = {2015},
  publisher = {Elsevier}
}

@book{stiff2016persuasive,
  title     = {Persuasive communication},
  author    = {Stiff, James B and Mongeau, Paul A},
  year      = {2016},
  publisher = {Guilford Publications}
}

@misc{carrasco-farreLargeLanguageModels2024,
  title      = {Large {Language} {Models} are as persuasive as humans, but how? {About} the cognitive effort and moral-emotional language of {LLM} arguments},
  shorttitle = {Large {Language} {Models} are as persuasive as humans, but how?},
  url        = {http://arxiv.org/abs/2404.09329},
  doi        = {10.48550/arXiv.2404.09329},
  urldate    = {2024-06-10},
  publisher  = {arXiv},
  author     = {Carrasco-Farre, Carlos},
  month      = apr,
  year       = {2024},
  note       = {arXiv:2404.09329 [cs]}
}

@article{breumPersuasivePowerLarge2024,
  title     = {The {Persuasive} {Power} of {Large} {Language} {Models}},
  volume    = {18},
  copyright = {Copyright (c) 2024 Association for the Advancement of Artificial Intelligence},
  issn      = {2334-0770},
  url       = {https://ojs.aaai.org/index.php/ICWSM/article/view/31304},
  doi       = {10.1609/icwsm.v18i1.31304},
  language  = {en},
  urldate   = {2024-06-10},
  journal   = {Proceedings of the International AAAI Conference on Web and Social Media},
  author    = {Breum, Simon Martin and Egdal, Daniel Vædele and Mortensen, Victor Gram and Møller, Anders Giovanni and Aiello, Luca Maria},
  month     = may,
  year      = {2024},
  pages     = {152--163}
}

@misc{liQuantifyingImpactLarge2024,
  address  = {Rochester, NY},
  type     = {{SSRN} {Scholarly} {Paper}},
  title    = {Quantifying the {Impact} of {Large} {Language} {Models} on {Collective} {Opinion} {Dynamics}},
  url      = {https://papers.ssrn.com/abstract=4688547},
  doi      = {10.2139/ssrn.4688547},
  language = {en},
  urldate  = {2024-06-14},
  author   = {Li, Chao and Su, Xing and Han, Haoying and Xue, Cong and Zheng, Chunmo and Fan, Chao},
  month    = jan,
  year     = {2024}
}

@misc{hackenburgComparingPersuasivenessRoleplaying2023,
  title     = {Comparing the persuasiveness of role-playing large language models and human experts on polarized {U}.{S}. political issues},
  url       = {https://osf.io/ey8db},
  doi       = {10.31219/osf.io/ey8db},
  language  = {en-us},
  urldate   = {2024-06-10},
  publisher = {OSF},
  author    = {Hackenburg, Kobi and Ibrahim, Lujain and Tappin, Ben M. and Tsakiris, Manos},
  month     = dec,
  year      = {2023}
}

@misc{costelloDurablyReducingConspiracy2024,
  title     = {Durably reducing conspiracy beliefs through dialogues with {AI}},
  url       = {https://osf.io/xcwdn},
  doi       = {10.31234/osf.io/xcwdn},
  language  = {en-us},
  urldate   = {2024-06-10},
  publisher = {OSF},
  author    = {Costello, Thomas H. and Pennycook, Gordon and Rand, David},
  month     = apr,
  year      = {2024}
}

@article{bohmPeopleDevalueGenerative2023,
  title     = {People devalue generative {AI}’s competence but not its advice in addressing societal and personal challenges},
  volume    = {1},
  copyright = {2023 The Author(s)},
  issn      = {2731-9121},
  url       = {https://www.nature.com/articles/s44271-023-00032-x},
  doi       = {10.1038/s44271-023-00032-x},
  language  = {en},
  number    = {1},
  urldate   = {2024-06-14},
  journal   = {Communications Psychology},
  author    = {Böhm, Robert and Jörling, Moritz and Reiter, Leonhard and Fuchs, Christoph},
  month     = nov,
  year      = {2023},
  note      = {Publisher: Nature Publishing Group},
  pages     = {1--10}
}

@article{matzPotentialGenerativeAI2024,
  title     = {The potential of generative {AI} for personalized persuasion at scale},
  volume    = {14},
  copyright = {2024 The Author(s)},
  issn      = {2045-2322},
  url       = {https://www.nature.com/articles/s41598-024-53755-0},
  doi       = {10.1038/s41598-024-53755-0},
  language  = {en},
  number    = {1},
  urldate   = {2024-06-10},
  journal   = {Scientific Reports},
  author    = {Matz, S. C. and Teeny, J. D. and Vaid, S. S. and Peters, H. and Harari, G. M. and Cerf, M.},
  month     = feb,
  year      = {2024},
  note      = {Publisher: Nature Publishing Group},
  pages     = {4692}
}

@article{goldsteinHowPersuasiveAIgenerated2024,
  title    = {How persuasive is {AI}-generated propaganda?},
  volume   = {3},
  issn     = {2752-6542},
  url      = {https://doi.org/10.1093/pnasnexus/pgae034},
  doi      = {10.1093/pnasnexus/pgae034},
  number   = {2},
  urldate  = {2024-06-10},
  journal  = {PNAS Nexus},
  author   = {Goldstein, Josh A and Chao, Jason and Grossman, Shelby and Stamos, Alex and Tomz, Michael},
  month    = feb,
  year     = {2024},
  pages    = {pgae034}
}

@article{hackenburgEvaluatingPersuasiveInfluence2024,
  title    = {Evaluating the persuasive influence of political microtargeting with large language models},
  volume   = {121},
  url      = {https://www.pnas.org/doi/10.1073/pnas.2403116121},
  doi      = {10.1073/pnas.2403116121},
  number   = {24},
  urldate  = {2024-06-10},
  journal  = {Proceedings of the National Academy of Sciences},
  author   = {Hackenburg, Kobi and Margetts, Helen},
  month    = jun,
  year     = {2024},
  note     = {Publisher: Proceedings of the National Academy of Sciences},
  pages    = {e2403116121}
}

@article{baiArtificialIntelligenceCan2023,
  title    = {Artificial {Intelligence} {Can} {Persuade} {Humans} on {Political} {Issues}},
  url      = {https://www.researchsquare.com/article/rs-3238396/v1},
  doi      = {10.21203/rs.3.rs-3238396/v1},
  urldate  = {2024-06-10},
  author   = {Bai, Hui and Voelkel, Jan and Eichstaedt, Johannes and Willer, Robb},
  month    = sep,
  year     = {2023},
  note     = {ISSN: 2693-5015}
}

@article{simchonPersuasiveEffectsPolitical2024,
  title    = {The persuasive effects of political microtargeting in the age of generative artificial intelligence},
  volume   = {3},
  issn     = {2752-6542},
  url      = {https://doi.org/10.1093/pnasnexus/pgae035},
  doi      = {10.1093/pnasnexus/pgae035},
  number   = {2},
  urldate  = {2024-06-10},
  journal  = {PNAS Nexus},
  author   = {Simchon, Almog and Edwards, Matthew and Lewandowsky, Stephan},
  month    = feb,
  year     = {2024},
  pages    = {pgae035}
}

@article{teigenPersuasivenessArgumentsAIsource2024,
  title    = {Persuasiveness of arguments with {AI}-source labels},
  volume   = {46},
  url      = {https://escholarship.org/uc/item/6t82g70v},
  language = {en},
  number   = {0},
  urldate  = {2024-06-14},
  journal  = {Proceedings of the Annual Meeting of the Cognitive Science Society},
  author   = {Teigen, Cassandra and Madsen, Jens Koed and George, Nicole Lauren and Yousefi, Sayeh},
  year     = {2024}
}

@article{palmerLargeLanguageModels2023,
  title      = {Large {Language} {Models} {Can} {Argue} in {Convincing} {Ways} {About} {Politics}, {But} {Humans} {Dislike} {AI} {Authors}: implications for {Governance}},
  volume     = {75},
  issn       = {0032-3187, 2041-0611},
  shorttitle = {Large {Language} {Models} {Can} {Argue} in {Convincing} {Ways} {About} {Politics}, {But} {Humans} {Dislike} {AI} {Authors}},
  url        = {https://www.tandfonline.com/doi/full/10.1080/00323187.2024.2335471},
  doi        = {10.1080/00323187.2024.2335471},
  language   = {en},
  number     = {3},
  urldate    = {2024-06-14},
  journal    = {Political Science},
  author     = {Palmer, Alexis and Spirling, Arthur},
  month      = sep,
  year       = {2023},
  pages      = {281--291}
}

@misc{durmus2024persuasion,
  title    = {Measuring the persuasiveness of language models},
  url      = {https://www.anthropic.com/news/measuring-model-persuasiveness},
  author   = {Durmus, Esin and Lovitt, Liane and Tamkin, Alex and Ritchie, Stuart and Clark, Jack and Ganguli, Deep},
  month    = apr,
  year     = {2024}
}

@article{chenWouldAIChatbot2023,
  title      = {Would an {AI} chatbot persuade you: an empirical answer from the elaboration likelihood model},
  volume     = {ahead-of-print},
  issn       = {0959-3845},
  shorttitle = {Would an {AI} chatbot persuade you},
  url        = {https://doi.org/10.1108/ITP-10-2021-0764},
  doi        = {10.1108/ITP-10-2021-0764},
  number     = {ahead-of-print},
  urldate    = {2024-06-14},
  journal    = {Information Technology \& People},
  author     = {Chen, Qian and Yin, Changqin and Gong, Yeming},
  month      = jan,
  year       = {2023}
}

@misc{pauliMeasuringBenchmarkingLarge2024,
  title     = {Measuring and {Benchmarking} {Large} {Language} {Models}' {Capabilities} to {Generate} {Persuasive} {Language}},
  url       = {http://arxiv.org/abs/2406.17753},
  doi       = {10.48550/arXiv.2406.17753},
  urldate   = {2024-07-05},
  publisher = {arXiv},
  author    = {Pauli, Amalie Brogaard and Augenstein, Isabelle and Assent, Ira},
  month     = jun,
  year      = {2024},
  note      = {arXiv:2406.17753 [cs]}
}

@misc{hackenburgEvidenceLogScaling2024,
  title     = {Evidence of a log scaling law for political persuasion with large language models},
  url       = {http://arxiv.org/abs/2406.14508},
  doi       = {10.48550/arXiv.2406.14508},
  urldate   = {2024-07-05},
  publisher = {arXiv},
  author    = {Hackenburg, Kobi and Tappin, Ben M. and Röttger, Paul and Hale, Scott and Bright, Jonathan and Margetts, Helen},
  month     = jun,
  year      = {2024},
  note      = {arXiv:2406.14508 [cs]}
}

@inproceedings{meguellatiHowGoodAre2024,
  address   = {New York, NY, USA},
  series    = {{WWW} '24},
  title     = {How {Good} are {LLMs} in {Generating} {Personalized} {Advertisements}?},
  isbn      = {9798400701726},
  url       = {https://doi.org/10.1145/3589335.3651520},
  doi       = {10.1145/3589335.3651520},
  urldate   = {2024-07-05},
  booktitle = {Companion {Proceedings} of the {ACM} on {Web} {Conference} 2024},
  publisher = {Association for Computing Machinery},
  author    = {Meguellati, Elyas and Han, Lei and Bernstein, Abraham and Sadiq, Shazia and Demartini, Gianluca},
  month     = may,
  year      = {2024},
  pages     = {826--829}
}

@article{shin2025adoption,
  title={The adoption and efficacy of large language models: Evidence from consumer complaints in the financial industry},
  author={Shin, Minkyu and Kim, Jin and Shin, Jiwoong},
  journal={Available at SSRN 5004194},
  year={2025}
}

@article{karinshakWorkingAIPersuade2023a,
  title      = {Working {With} {AI} to {Persuade}: {Examining} a {Large} {Language} {Model}'s {Ability} to {Generate} {Pro}-{Vaccination} {Messages}},
  volume     = {7},
  issn       = {2573-0142},
  shorttitle = {Working {With} {AI} to {Persuade}},
  url        = {https://dl.acm.org/doi/10.1145/3579592},
  doi        = {10.1145/3579592},
  language   = {en},
  number     = {CSCW1},
  urldate    = {2024-07-05},
  journal    = {Proceedings of the ACM on Human-Computer Interaction},
  author     = {Karinshak, Elise and Liu, Sunny Xun and Park, Joon Sung and Hancock, Jeffrey T.},
  month      = apr,
  year       = {2023},
  pages      = {1--29}
}

@misc{furumaiZeroshotPersuasiveChatbots2024,
  title     = {Zero-shot {Persuasive} {Chatbots} with {LLM}-{Generated} {Strategies} and {Information} {Retrieval}},
  url       = {http://arxiv.org/abs/2407.03585},
  doi       = {10.48550/arXiv.2407.03585},
  urldate   = {2024-07-10},
  publisher = {arXiv},
  author    = {Furumai, Kazuaki and Legaspi, Roberto and Vizcarra, Julio and Yamazaki, Yudai and Nishimura, Yasutaka and Semnani, Sina J. and Ikeda, Kazushi and Shi, Weiyan and Lam, Monica S.},
  month     = jul,
  year      = {2024},
  note      = {arXiv:2407.03585 [cs]}
}

@misc{yoonDesigningEvaluatingMultiChatbot2024,
  title      = {Designing and {Evaluating} {Multi}-{Chatbot} {Interface} for {Human}-{AI} {Communication}: {Preliminary} {Findings} from a {Persuasion} {Task}},
  shorttitle = {Designing and {Evaluating} {Multi}-{Chatbot} {Interface} for {Human}-{AI} {Communication}},
  url        = {http://arxiv.org/abs/2406.19648},
  doi        = {10.48550/arXiv.2406.19648},
  urldate    = {2024-07-10},
  publisher  = {arXiv},
  author     = {Yoon, Sion and Kim, Tae Eun and Oh, Yoo Jung},
  month      = jun,
  year       = {2024},
  note       = {arXiv:2406.19648 [cs]}
}

@inproceedings{metzgerEmpoweringCalibratedDis2024,
  address    = {New York, NY, USA},
  series     = {{CHI} '24},
  title      = {Empowering {Calibrated} ({Dis}-){Trust} in {Conversational} {Agents}: {A} {User} {Study} on the {Persuasive} {Power} of {Limitation} {Disclaimers} vs. {Authoritative} {Style}},
  isbn       = {9798400703300},
  shorttitle = {Empowering {Calibrated} ({Dis}-){Trust} in {Conversational} {Agents}},
  url        = {https://doi.org/10.1145/3613904.3642122},
  doi        = {10.1145/3613904.3642122},
  urldate    = {2024-07-10},
  booktitle  = {Proceedings of the {CHI} {Conference} on {Human} {Factors} in {Computing} {Systems}},
  publisher  = {Association for Computing Machinery},
  author     = {Metzger, Luise and Miller, Linda and Baumann, Martin and Kraus, Johannes},
  month      = may,
  year       = {2024},
  pages      = {1--19}
}

@misc{chenCombatingMisinformationAge2023,
  title      = {Combating {Misinformation} in the {Age} of {LLMs}: {Opportunities} and {Challenges}},
  shorttitle = {Combating {Misinformation} in the {Age} of {LLMs}},
  url        = {http://arxiv.org/abs/2311.05656},
  doi        = {10.48550/arXiv.2311.05656},
  urldate    = {2024-06-20},
  publisher  = {arXiv},
  author     = {Chen, Canyu and Shu, Kai},
  month      = nov,
  year       = {2023},
  note       = {arXiv:2311.05656 [cs]}
}

@misc{carrollCharacterizingManipulationAI2023a,
  title     = {Characterizing {Manipulation} from {AI} {Systems}},
  url       = {http://arxiv.org/abs/2303.09387},
  doi       = {10.48550/arXiv.2303.09387},
  urldate   = {2024-09-18},
  publisher = {arXiv},
  author    = {Carroll, Micah and Chan, Alan and Ashton, Henry and Krueger, David},
  month     = oct,
  year      = {2023},
  note      = {arXiv:2303.09387 [cs]}
}

@misc{hendrycksOverviewCatastrophicAI2023a,
  title     = {An {Overview} of {Catastrophic} {AI} {Risks}},
  url       = {http://arxiv.org/abs/2306.12001},
  doi       = {10.48550/arXiv.2306.12001},
  urldate   = {2024-09-18},
  publisher = {arXiv},
  author    = {Hendrycks, Dan and Mazeika, Mantas and Woodside, Thomas},
  month     = oct,
  year      = {2023},
  note      = {arXiv:2306.12001 [cs]}
}

@misc{buyl2024largelanguagemodelsreflect,
  title         = {Large Language Models Reflect the Ideology of their Creators},
  author        = {Maarten Buyl and Alexander Rogiers and Sander Noels and 
                   Iris Dominguez-Catena and Edith Heiter and Raphael Romero and 
                   Iman Johary and Alexandru-Cristian Mara and Jefrey Lijffijt and 
                   Tijl De Bie},
  year          = {2024},
  eprint        = {2410.18417},
  archiveprefix = {arXiv},
  primaryclass  = {cs.CL},
  url           = {https://arxiv.org/abs/2410.18417},
  doi           = {10.48550/arXiv.2410.18417},
  note          = {arXiv:2410.18417 [cs]}
}

@inproceedings{zhouSyntheticLiesUnderstanding2023,
  address    = {New York, NY, USA},
  series     = {{CHI} '23},
  title      = {Synthetic {Lies}: {Understanding} {AI}-{Generated} {Misinformation} and {Evaluating} {Algorithmic} and {Human} {Solutions}},
  isbn       = {978-1-4503-9421-5},
  shorttitle = {Synthetic {Lies}},
  url        = {https://dl.acm.org/doi/10.1145/3544548.3581318},
  doi        = {10.1145/3544548.3581318},
  urldate    = {2024-06-14},
  booktitle  = {Proceedings of the 2023 {CHI} {Conference} on {Human} {Factors} in {Computing} {Systems}},
  publisher  = {Association for Computing Machinery},
  author     = {Zhou, Jiawei and Zhang, Yixuan and Luo, Qianni and Parker, Andrea G and De Choudhury, Munmun},
  month      = apr,
  year       = {2023},
  pages      = {1--20}
}

@article{spitaleAIModelGPT32023,
  title    = {{AI} model {GPT}-3 (dis)informs us better than humans},
  volume   = {9},
  url      = {https://www.science.org/doi/10.1126/sciadv.adh1850},
  doi      = {10.1126/sciadv.adh1850},
  number   = {26},
  urldate  = {2024-06-14},
  journal  = {Science Advances},
  author   = {Spitale, Giovanni and Biller-Andorno, Nikola and Germani, Federico},
  month    = jun,
  year     = {2023},
  note     = {Publisher: American Association for the Advancement of Science},
  pages    = {eadh1850}
}

@article{limEffectSourceDisclosure2024,
  title      = {The effect of source disclosure on evaluation of {AI}-generated messages: {A} two-part study},
  volume     = {2},
  issn       = {29498821},
  shorttitle = {The effect of source disclosure on evaluation of {AI}-generated messages},
  url        = {http://arxiv.org/abs/2311.15544},
  doi        = {10.1016/j.chbah.2024.100058},
  language   = {en},
  number     = {1},
  urldate    = {2024-06-21},
  journal    = {Computers in Human Behavior: Artificial Humans},
  author     = {Lim, Sue and Schmälzle, Ralf},
  month      = jan,
  year       = {2024},
  note       = {arXiv:2311.15544 [cs]},
  pages      = {100058}
}

@article{zhangHumanFavoritismNot2023,
  title      = {Human favoritism, not {AI} aversion: {People}’s perceptions (and bias) toward generative {AI}, human experts, and human–{GAI} collaboration in persuasive content generation},
  volume     = {18},
  issn       = {1930-2975},
  shorttitle = {Human favoritism, not {AI} aversion},
  url        = {https://www.cambridge.org/core/journals/judgment-and-decision-making/article/human-favoritism-not-ai-aversion-peoples-perceptions-and-bias-toward-generative-ai-human-experts-and-humangai-collaboration-in-persuasive-content-generation/419C4BD9CE82673EAF1D8F6C350C4FA8},
  doi        = {10.1017/jdm.2023.37},
  language   = {en},
  urldate    = {2024-06-13},
  journal    = {Judgment and Decision Making},
  author     = {Zhang, Yunhao and Gosline, Renée},
  month      = jan,
  year       = {2023},
  pages      = {e41}
}

@article{salvi2025conversational,
  title     = {On the conversational persuasiveness of GPT-4},
  author    = {Salvi, Francesco and Horta Ribeiro, Manoel and Gallotti, Riccardo and West, Robert},
  journal   = {Nature Human Behaviour},
  volume = {9},
  pages     = {1645--1653},
  year      = {2025},
  publisher = {Nature Publishing Group UK London}
}

@article{hackenburg2025scaling,
  title     = {Scaling language model size yields diminishing returns for single-message political persuasion},
  author    = {Hackenburg, Kobi and Tappin, Ben M and R{\"o}ttger, Paul and Hale, Scott A and Bright, Jonathan and Margetts, Helen},
  journal   = {Proceedings of the National Academy of Sciences},
  volume    = {122},
  number    = {10},
  pages     = {e2413443122},
  year      = {2025},
  publisher = {National Academy of Sciences}
}

@article{bai2025llm,
  author            = {Bai, Hui and Voelkel, Jan G. and Muldowney, Shane and Eichstaedt, Johannes C. and Willer, Robb},
  title             = {LLM-generated messages can persuade humans on policy issues},
  year              = {2025},
  journal           = {Nature Communications },
  volume            = {16},
  number            = {1},
  doi               = {10.1038/s41467-025-61345-5},
  url               = {https://www.scopus.com/inward/record.uri?eid=2-s2.0-105009546258&doi=10.1038%2fs41467-025-61345-5&partnerID=40&md5=c865532ec84f719e0d2f1f3eecda25e3},
  type              = {Article},
  publication_stage = {Final},
  source            = {Scopus},
  note              = {Cited by: 3; All Open Access, Gold Open Access, Green Open Access}
}

@article{he2025enhancing,
  title    = {Enhancing belief consistency of large language model agents in decision-making process based on attribution theory},
  journal  = {Expert Systems with Applications},
  volume   = {297},
  pages    = {129273},
  year     = {2026},
  issn     = {0957-4174},
  doi      = {https://doi.org/10.1016/j.eswa.2025.129273},
  url      = {https://www.sciencedirect.com/science/article/pii/S0957417425028891},
  author   = {Guoxiu He and Meicong Zhang and Tiancheng Su and Li Ma and Xiaomin Zhu}
}

@article{sun2026cutting,
  title    = {When cutting edge meets silver tongue: Understanding the word-of-machine effect on travel decisions},
  journal  = {Tourism Management},
  volume   = {112},
  pages    = {105271},
  year     = {2026},
  issn     = {0261-5177},
  doi      = {https://doi.org/10.1016/j.tourman.2025.105271},
  url      = {https://www.sciencedirect.com/science/article/pii/S0261517725001414},
  author   = {Danni Sun and IpKin Anthony Wong and Xiling Xiong and Shina Li}
}

@inproceedings{coppolillo2025engagement,
  title     = {Engagement-driven content generation with large language models},
  author    = {Coppolillo, Erica and Cinus, Federico and Minici, Marco and Bonchi, Francesco and Manco, Giuseppe},
  booktitle = {Proceedings of the 31st ACM SIGKDD Conference on Knowledge Discovery and Data Mining V. 2},
  pages     = {369--379},
  year      = {2025}
}

@inproceedings{doudkin2025synthetic,
  title     = {From Synthetic to Human: The Gap Between AI-Predicted and Actual Pro-Environmental Behavior Change After Chatbot Persuasion},
  author    = {Doudkin, Alexander and Pataranutaporn, Pat and Maes, Pattie},
  booktitle = {Proceedings of the 7th ACM Conference on Conversational User Interfaces},
  pages     = {1--18},
  year      = {2025}
}

@inproceedings{meguellati2025duality,
  title     = {The Duality of Persuasion: Between Personalization and Detection},
  author    = {Meguellati, Elyas},
  booktitle = {Companion Proceedings of the ACM on Web Conference 2025},
  pages     = {705--708},
  year      = {2025}
}

@article{argyle2025testing,
  title     = {Testing theories of political persuasion using AI},
  author    = {Argyle, Lisa P and Busby, Ethan C and Gubler, Joshua R and Lyman, Alex and Olcott, Justin and Pond, Jackson and Wingate, David},
  journal   = {Proceedings of the National Academy of Sciences},
  volume    = {122},
  number    = {18},
  pages     = {e2412815122},
  year      = {2025},
  publisher = {National Academy of Sciences}
}

@inproceedings{biswas2025mind,
  author            = {Biswas, Shreyan and Erlei, Alexander and Gadiraju, Ujwal},
  title             = {Mind the Gap! Choice Independence in Using Multilingual LLMs for Persuasive Co-Writing Tasks in Different Languages},
  year              = {2025},
  journal           = {Conference on Human Factors in Computing Systems - Proceedings },
  doi               = {10.1145/3706598.3713201},
  url               = {https://www.scopus.com/inward/record.uri?eid=2-s2.0-105005751330&doi=10.1145%2f3706598.3713201&partnerID=40&md5=48a8ad12f62339ae075222253b34cdb6},
  type              = {Conference paper},
  publication_stage = {Final},
  source            = {Scopus},
  note              = {Cited by: 5; All Open Access, Gold Open Access}
}

@inproceedings{ataguba2025persuasion,
  title        = {Persuasion and Behavior Change in ChatGPT-Based Dietary Management},
  author       = {Ataguba, Grace and Oyebode, Oladapo and Orji, Fidelia and Henry, Kosi Clinton and Orji, Rita},
  booktitle    = {2025 IEEE Conference on Serious Games and Applications for Health (SeGAH)},
  pages        = {1--8},
  year         = {2025},
  organization = {IEEE}
}

@inproceedings{nezhad2025adaptive,
  author    = {Nezhad, Mansoureh Motahari and Kisomi, Maysam Avakh and Gholinezhad, Fatemeh},
  booktitle = {2025 11th International Conference on Web Research (ICWR)},
  title     = {Adaptive Persuasion in Conversational AI: An LLM-Driven Framework for Dynamic Strategy Switching via Personality and Sentiment Analysis},
  year      = {2025},
  volume    = {},
  number    = {},
  pages     = {145-149},
  doi       = {10.1109/ICWR65219.2025.11006192}
}

@inproceedings{kong2025huper,
  author    = {Kong, Lingzhen
               and Jin, Chuhao
               and Song, Ruihua
               and Wang, Xiting
               and Chen, Yu},
  editor    = {Yuan, Shuhan
               and Malliaros, Fragkiskos
               and Zheng, Xin},
  title     = {HuPer: Human Factors Integrated Persuasive Dialogue Models},
  booktitle = {Trends and Applications in Knowledge Discovery and Data Mining},
  year      = {2025},
  publisher = {Springer Nature Singapore},
  address   = {Singapore},
  pages     = {298--310},
  isbn      = {978-981-96-8197-6}
}

@inproceedings{elaraby2024persuasiveness,
  title     = {Persuasiveness of Generated Free-Text Rationales in Subjective Decisions: A Case Study on Pairwise Argument Ranking},
  author    = {Elaraby, Mohamed  and
               Litman, Diane  and
               Li, Xiang Lorraine  and
               Magooda, Ahmed},
  editor    = {Al-Onaizan, Yaser  and
               Bansal, Mohit  and
               Chen, Yun-Nung},
  booktitle = {Findings of the Association for Computational Linguistics: EMNLP 2024},
  month     = nov,
  year      = {2024},
  address   = {Miami, Florida, USA},
  publisher = {Association for Computational Linguistics},
  url       = {https://aclanthology.org/2024.findings-emnlp.836/},
  doi       = {10.18653/v1/2024.findings-emnlp.836},
  pages     = {14311--14329}
}

@inproceedings{jin2024persuading,
  title     = {Persuading across Diverse Domains: a Dataset and Persuasion Large Language Model},
  author    = {Jin, Chuhao  and
               Ren, Kening  and
               Kong, Lingzhen  and
               Wang, Xiting  and
               Song, Ruihua  and
               Chen, Huan},
  editor    = {Ku, Lun-Wei  and
               Martins, Andre  and
               Srikumar, Vivek},
  booktitle = {Proceedings of the 62nd Annual Meeting of the Association for Computational Linguistics (Volume 1: Long Papers)},
  month     = aug,
  year      = {2024},
  address   = {Bangkok, Thailand},
  publisher = {Association for Computational Linguistics},
  url       = {https://aclanthology.org/2024.acl-long.92/},
  doi       = {10.18653/v1/2024.acl-long.92},
  pages     = {1678--1706}
}

@inproceedings{wu2024mindshift,
  author    = {Wu, Ruolan and Yu, Chun and Pan, Xiaole and Liu, Yujia and Zhang, Ningning and Fu, Yue and Wang, Yuhan and Zheng, Zhi and Chen, Li and Jiang, Qiaolei and Xu, Xuhai and Shi, Yuanchun},
  title     = {MindShift: Leveraging Large Language Models for Mental-States-Based Problematic Smartphone Use Intervention},
  year      = {2024},
  isbn      = {9798400703300},
  publisher = {Association for Computing Machinery},
  address   = {New York, NY, USA},
  url       = {https://doi.org/10.1145/3613904.3642790},
  doi       = {10.1145/3613904.3642790},
  booktitle = {Proceedings of the 2024 CHI Conference on Human Factors in Computing Systems},
  articleno = {248},
  numpages  = {24},
  location  = {Honolulu, HI, USA},
  series    = {CHI '24}
}

@inproceedings{qin2024beyond,
  title     = {Beyond Persuasion: Towards Conversational Recommender System with Credible Explanations},
  author    = {Qin, Peixin  and
               Huang, Chen  and
               Deng, Yang  and
               Lei, Wenqiang  and
               Chua, Tat-Seng},
  editor    = {Al-Onaizan, Yaser  and
               Bansal, Mohit  and
               Chen, Yun-Nung},
  booktitle = {Findings of the Association for Computational Linguistics: EMNLP 2024},
  month     = nov,
  year      = {2024},
  address   = {Miami, Florida, USA},
  publisher = {Association for Computational Linguistics},
  url       = {https://aclanthology.org/2024.findings-emnlp.247/},
  doi       = {10.18653/v1/2024.findings-emnlp.247},
  pages     = {4264--4282}
}

@inproceedings{calle2024towards,
  author    = {Calle, Paul and Shao, Ruosi and Liu, Yunlong and H\'{e}bert, Emily T and Kendzor, Darla and Neil, Jordan and Businelle, Michael and Pan, Chongle},
  title     = {Towards AI-Driven Healthcare: Systematic Optimization, Linguistic Analysis, and Clinicians’ Evaluation of Large Language Models for Smoking Cessation Interventions},
  year      = {2024},
  isbn      = {9798400703300},
  publisher = {Association for Computing Machinery},
  address   = {New York, NY, USA},
  url       = {https://doi.org/10.1145/3613904.3641965},
  doi       = {10.1145/3613904.3641965},
  booktitle = {Proceedings of the 2024 CHI Conference on Human Factors in Computing Systems},
  articleno = {436},
  numpages  = {16},
  location  = {Honolulu, HI, USA},
  series    = {CHI '24}
}

@inproceedings{el2024improving,
  title     = {Improving Argument Effectiveness Across Ideologies using Instruction-tuned Large Language Models},
  author    = {El Baff, Roxanne  and
               Khatib, Khalid Al  and
               Alshomary, Milad  and
               Konen, Kai  and
               Stein, Benno  and
               Wachsmuth, Henning},
  editor    = {Al-Onaizan, Yaser  and
               Bansal, Mohit  and
               Chen, Yun-Nung},
  booktitle = {Findings of the Association for Computational Linguistics: EMNLP 2024},
  month     = nov,
  year      = {2024},
  address   = {Miami, Florida, USA},
  publisher = {Association for Computational Linguistics},
  url       = {https://aclanthology.org/2024.findings-emnlp.265/},
  doi       = {10.18653/v1/2024.findings-emnlp.265},
  pages     = {4604--4622}
}

@article{elhissoufi2024leveraging,
  title   = {Leveraging Generative Large Language Models for Optimizing Sales Arguments Creation: An Evaluation of GPT-4 Capabilities},
  author  = {Elhissoufi, Mustapha and Nfaoui, El Habib and Alla, Lhoussaine and Elghalfiki, Jawad},
  journal = {International Journal of Intelligent Engineering \& Systems},
  volume  = {17},
  number  = {5},
  year    = {2024}
}

@article{xia2025comparison,
  title     = {A comparison of the persuasiveness of human and ChatGPT generated pro-vaccine messages for HPV},
  author    = {Xia, Dengke and Song, Mengyao and Zhu, Tingshao},
  journal   = {Frontiers in Public Health},
  volume    = {12},
  pages     = {1515871},
  year      = {2025},
  publisher = {Frontiers Media SA}
}

@incollection{wilczynski2024resistance,
  title     = {Resistance against manipulative ai: key factors and possible actions},
  author    = {Wilczy{\'n}ski, Piotr and Mieleszczenko-Kowszewicz, Wiktoria and Biecek, Przemys{\l}aw},
  booktitle = {ECAI 2024},
  pages     = {802--809},
  year      = {2024},
  publisher = {IOS Press}
}

@inproceedings{fetrati2025leveraging,
  author    = {Fetrati, Hemad and Chan, Gerry and Orji, Rita},
  title     = {Leveraging Generative and Rule-Based Models for Persuasive STI Education: A Multi-Chatbot Mobile Application},
  year      = {2025},
  isbn      = {9798400715273},
  publisher = {Association for Computing Machinery},
  address   = {New York, NY, USA},
  url       = {https://doi.org/10.1145/3719160.3737619},
  doi       = {10.1145/3719160.3737619},
  booktitle = {Proceedings of the 7th ACM Conference on Conversational User Interfaces},
  articleno = {16},
  numpages  = {9},
  location  = {
               },
  series    = {CUI '25}
}

@inproceedings{vahidov2025customer,
  author = {Vahidov, Rustam and Carbonneau, Réal},
  year   = {2025},
  month  = {01},
  pages  = {},
  title  = {Customer – Software Agent Negotiations Using Large Language Model: An Experimental Study},
  doi    = {10.24251/HICSS.2025.522}
}

@inproceedings{schneiders2025objection,
  author    = {Schneiders, Eike and Seabrooke, Tina and Krook, Joshua and Hyde, Richard and Leesakul, Natalie and Clos, Jeremie and Fischer, Joel E},
  title     = {Objection Overruled! Lay People can Distinguish Large Language Models from Lawyers, but still Favour Advice from an LLM},
  year      = {2025},
  isbn      = {9798400713941},
  publisher = {Association for Computing Machinery},
  address   = {New York, NY, USA},
  url       = {https://doi.org/10.1145/3706598.3713470},
  doi       = {10.1145/3706598.3713470},
  booktitle = {Proceedings of the 2025 CHI Conference on Human Factors in Computing Systems},
  articleno = {1201},
  numpages  = {14},
  location  = {
               },
  series    = {CHI '25}
}

@inproceedings{saenger2024autopersuade,
  title     = {{A}uto{P}ersuade: A Framework for Evaluating and Explaining Persuasive Arguments},
  author    = {Saenger, Till Raphael  and
               Hinck, Musashi  and
               Grimmer, Justin  and
               Stewart, Brandon M.},
  editor    = {Al-Onaizan, Yaser  and
               Bansal, Mohit  and
               Chen, Yun-Nung},
  booktitle = {Proceedings of the 2024 Conference on Empirical Methods in Natural Language Processing},
  month     = nov,
  year      = {2024},
  address   = {Miami, Florida, USA},
  publisher = {Association for Computational Linguistics},
  url       = {https://aclanthology.org/2024.emnlp-main.913/},
  doi       = {10.18653/v1/2024.emnlp-main.913},
  pages     = {16325--16342}
}

@inproceedings{corro2025exploring,
  title        = {Exploring the Potential and Limitations of Large Language Models to Control the Behavior of Embodied Persuasive Agents},
  author       = {Corr{\`o}, Christian and Chittaro, Luca},
  booktitle    = {International Conference on Persuasive Technology},
  pages        = {61--73},
  year         = {2025},
  organization = {Springer}
}

@inproceedings{cima2025contextualized,
  title     = {Contextualized counterspeech: Strategies for adaptation, personalization, and evaluation},
  author    = {Cima, Lorenzo and Miaschi, Alessio and Trujillo, Amaury and Avvenuti, Marco and Dell'Orletta, Felice and Cresci, Stefano},
  booktitle = {Proceedings of the ACM on Web Conference 2025},
  pages     = {5022--5033},
  year      = {2025}
}

@article{karakacs2025changes,
  title     = {Changes in attitudes toward meat consumption after chatting with a large language model},
  author    = {Karaka{\c{s}}, Neslihan and Jaeger, Bastian},
  journal   = {Social Influence},
  volume    = {20},
  number    = {1},
  pages     = {2475802},
  year      = {2025},
  publisher = {Taylor \& Francis}
}

@article{lim2023artificial,
  title     = {Artificial intelligence for health message generation: an empirical study using a large language model (LLM) and prompt engineering},
  author    = {Lim, Sue and Schm{\"a}lzle, Ralf},
  journal   = {Frontiers in Communication},
  volume    = {8},
  pages     = {1129082},
  year      = {2023},
  publisher = {Frontiers Media SA}
}

@article{dash2025persuasive,
  title     = {The persuasive potential of AI-paraphrased information at scale},
  author    = {Dash, Saloni and Xu, Yiwei and Jalbert, Madeline and Spiro, Emma S},
  journal   = {PNAS nexus},
  volume    = {4},
  number    = {7},
  pages     = {pgaf207},
  year      = {2025},
  publisher = {Oxford University Press US}
}

@article{kumar2025large,
  title     = {Large language model agents for improving engagement with behavior change interventions: Application to digital mindfulness},
  author    = {Kumar, Harsh and Yoo, Suhyeon and Zavaleta Bernuy, Angela and Shi, Jiakai and Luo, Huayin and Williams, Joseph Jay and Kuzminykh, Anastasia and Anderson, Ashton and Kornfield, Rachel},
  journal   = {Proceedings of the ACM on Human-Computer Interaction},
  volume    = {9},
  number    = {7},
  pages     = {1--44},
  year      = {2025},
  publisher = {ACM New York, NY, USA}
}

@inproceedings{potter2024hidden,
  title     = {Hidden Persuaders: {LLM}s' Political Leaning and Their Influence on Voters},
  author    = {Potter, Yujin  and
               Lai, Shiyang  and
               Kim, Junsol  and
               Evans, James  and
               Song, Dawn},
  editor    = {Al-Onaizan, Yaser  and
               Bansal, Mohit  and
               Chen, Yun-Nung},
  booktitle = {Proceedings of the 2024 Conference on Empirical Methods in Natural Language Processing},
  month     = nov,
  year      = {2024},
  address   = {Miami, Florida, USA},
  publisher = {Association for Computational Linguistics},
  url       = {https://aclanthology.org/2024.emnlp-main.244/},
  doi       = {10.18653/v1/2024.emnlp-main.244},
  pages     = {4244--4275}
}

@article{wang2019persuasion,
  title   = {Persuasion for good: Towards a personalized persuasive dialogue system for social good},
  author  = {Wang, Xuewei and Shi, Weiyan and Kim, Richard and Oh, Yoojung and Yang, Sijia and Zhang, Jingwen and Yu, Zhou},
  journal = {arXiv preprint arXiv:1906.06725},
  year    = {2019}
}

@article{hackenburg2025levers,
  author   = {Kobi Hackenburg  and Ben M. Tappin  and Luke Hewitt  and Ed Saunders  and Sid Black  and Hause Lin  and Catherine Fist  and Helen Margetts  and David G. Rand  and Christopher Summerfield },
  title    = {The levers of political persuasion with conversational artificial intelligence},
  journal  = {Science},
  volume   = {390},
  number   = {6777},
  pages    = {eaea3884},
  year     = {2025},
  doi      = {10.1126/science.aea3884},
  url      = {https://www.science.org/doi/abs/10.1126/science.aea3884},
  eprint   = {https://www.science.org/doi/pdf/10.1126/science.aea3884}
}

@article{lin2025persuading,
  title     = {Persuading voters using human--artificial intelligence dialogues},
  author    = {Lin, Hause and Czarnek, Gabriela and Lewis, Benjamin and White, Joshua P and Berinsky, Adam J and Costello, Thomas and Pennycook, Gordon and Rand, David G},
  journal   = {Nature},
  pages     = {1--8},
  year      = {2025},
  publisher = {Nature Publishing Group UK London}
}

@article{timm2025tailored,
  title   = {Tailored truths: Optimizing llm persuasion with personalization and fabricated statistics},
  author  = {Timm, Jasper and Talele, Chetan and Haimes, Jacob},
  journal = {arXiv preprint arXiv:2501.17273},
  year    = {2025}
}

@article{chen2025framework,
  title   = {A Framework to Assess the Persuasion Risks Large Language Model Chatbots Pose to Democratic Societies},
  author  = {Chen, Zhongren and Kalla, Joshua and Le, Quan and Nakamura-Sakai, Shinpei and Sekhon, Jasjeet and Wang, Ruixiao},
  journal = {arXiv preprint arXiv:2505.00036},
  year    = {2025}
}

@article{schoenegger2025large,
  title   = {Large Language Models Are More Persuasive Than Incentivized Human Persuaders},
  author  = {Schoenegger, Philipp and Salvi, Francesco and Liu, Jiacheng and Nan, Xiaoli and Debnath, Ramit and Fasolo, Barbara and Leivada, Evelina and Recchia, Gabriel and G{\"u}nther, Fritz and Zarifhonarvar, Ali and others},
  journal = {arXiv preprint arXiv:2505.09662},
  year    = {2025}
}

@article{liu2025llm,
  title   = {LLM can be a dangerous persuader: Empirical study of persuasion safety in large language models},
  author  = {Liu, Minqian and Xu, Zhiyang and Zhang, Xinyi and An, Heajun and Qadir, Sarvech and Zhang, Qi and Wisniewski, Pamela J and Cho, Jin-Hee and Lee, Sang Won and Jia, Ruoxi and others},
  journal = {arXiv preprint arXiv:2504.10430},
  year    = {2025}
}

@article{ramani2024persuasion,
  title   = {Persuasion games using large language models},
  author  = {Ramani, Ganesh Prasath and Karande, Shirish and Bhatia, Yash and others},
  journal = {arXiv preprint arXiv:2408.15879},
  year    = {2024}
}

@article{agarwal2025persuasion,
  title   = {When persuasion overrides truth in multi-agent llm debates: Introducing a confidence-weighted persuasion override rate (cw-por)},
  author  = {Agarwal, Mahak and Khanna, Divyam},
  journal = {arXiv preprint arXiv:2504.00374},
  year    = {2025}
}

@article{borah2025persuasion,
  title   = {Persuasion at Play: Understanding Misinformation Dynamics in Demographic-Aware Human-LLM Interactions},
  author  = {Borah, Angana and Mihalcea, Rada and P{\'e}rez-Rosas, Ver{\'o}nica},
  journal = {arXiv preprint arXiv:2503.02038},
  year    = {2025}
}

@article{zhou2025communication,
  title   = {Communication Styles and Reader Preferences of LLM and Human Experts in Explaining Health Information},
  author  = {Zhou, Jiawei and Venkatachalam, Kritika and Choi, Minje and Saha, Koustuv and De Choudhury, Munmun},
  journal = {arXiv preprint arXiv:2505.08143},
  year    = {2025}
}

@article{roy2025persuasiveness,
  title   = {Persuasiveness and Bias in LLM: Investigating the Impact of Persuasiveness and Reinforcement of Bias in Language Models},
  author  = {Roy, Saumya},
  journal = {arXiv preprint arXiv:2508.15798},
  year    = {2025}
}

@inproceedings{sasaki2025ai,
  title={When AI Gets Persuaded, Humans Follow: Inducing the Conformity Effect in Persuasive Dialogue},
  author={Sasaki, Rikuo and Inaba, Michimasa},
  booktitle={Proceedings of the 13th International Conference on Human-Agent Interaction},
  pages={1--9},
  year={2025}
}

@article{cheng2025towards,
  title   = {Towards Strategic Persuasion with Language Models},
  author  = {Cheng, Zirui and You, Jiaxuan},
  journal = {arXiv preprint arXiv:2509.22989},
  year    = {2025}
}

@article{havin2025can,
  title   = {Can (A) I Change Your Mind?},
  author  = {Havin, Miriam and Kleinman, Timna Wharton and Koren, Moran and Dover, Yaniv and Goldstein, Ariel},
  journal = {arXiv preprint arXiv:2503.01844},
  year    = {2025}
}

\end{document}